\documentclass[12pt]{article}
\usepackage[utf8]{inputenc}

\def\ourtitle{A Primer on Bayesian Neural Networks: \medskip
Review and Debates}

\usepackage{natbib}
\usepackage{amssymb}
\usepackage{amsfonts}
\usepackage{amsmath}
\usepackage{bm}
\usepackage{mathrsfs}
\usepackage{nicefrac}
\usepackage{mathtools}
\usepackage[margin=2.2cm]{geometry}

\usepackage{xcolor}
\definecolor{iblue}{RGB}{0,0,156}
\definecolor{ired}{RGB}{192,0,0}
\definecolor{igrey}{RGB}{108,123,139}

\usepackage[pagebackref=true]{hyperref}
\hypersetup{
	colorlinks=true,
	urlcolor=iblue,
	citecolor=iblue,
	linktoc=all,
	linkcolor=iblue
	}

\usepackage[capitalize]{cleveref}
	
\usepackage{tikz}
\usepackage[framemethod=TikZ]{mdframed}

\newcommand{\bx}{\boldsymbol{x}}

\newcommand{\bh}{\boldsymbol{h}}

\newcommand{\bw}{\boldsymbol{w}}

\newcommand{\by}{\boldsymbol{y}} 
\newcommand{\bg}{\boldsymbol{g}}

\newcommand{\ddr}{\mathrm{d}} 

\DeclareMathOperator*{\argmax}{arg\,max}
\DeclareMathOperator*{\argmin}{arg\,min}

\usepackage{amsthm}

\usepackage{tcolorbox}
\newtcolorbox{mybox}[3][]
{
  colframe = #2!25,
  colback  = #2!10,
  coltitle = #2!20!black,  
  title    = {#3},
  #1,
}

\usepackage{appendix}

\AtBeginEnvironment{subappendices}{%
\chapter*{Appendix}
\addcontentsline{toc}{chapter}{$\quad$ Appendix}
\counterwithin{figure}{section}
\counterwithin{table}{section}
}

\usepackage[acronym,nowarn]{glossaries-extra}
\glsdisablehyper
\makeglossaries

\newacronym[category=general,firstplural=Bayesian neural networks]{BNN}{BNN}{Bayesian neural network}
\newacronym[category=general,firstplural=deep neural networks]{DNN}{DNN}{deep neural network}
\newacronym[category=general,firstplural=Gaussian processes]{GP}{GP}{Gaussian process}
\newacronym[category=general]{CNN}{CNN}{convolutional neural network}
\newacronym[category=general]{RNN}{RNN}{recurrent neural network}
\newacronym[category=general]{NTK}{NTK}{neural tangent kernel}
\newacronym[category=general]{NN}{NN}{neural network}
\newacronym[category=special]{RELU}{ReLU}{rectified linear unit}
\newacronym[category=general]{MAP}{MAP}{maximum-a-posteriori}
\newacronym[category=general]{MLE}{MLE}{maximum likelihood estimation}
\newacronym[category=general]{MC}{MC}{Monte Carlo}
\newacronym[category=general]{MH}{MH}{Metropolis--Hastings}
\newacronym[category=general]{MCMC}{MCMC}{Markov chain Monte Carlo}
\newacronym[category=general]{MALA}{MALA}{Metropolis-Adjusted Langevin Algorithm}
\newacronym[category=general]{HMC}{HMC}{Hamiltonian Monte Carlo}
\newacronym[category=general]{NUTS}{NUTS}{No-U-Turn Sampler}
\newacronym[category=general]{VI}{VI}{variational inference}
\newacronym[category=general]{KL}{KL}{Kullback--Leibler}
\newacronym[category=general]{ELBO}{ELBO}{evidence lower bound}
\newacronym[category=general,firstplural=ordinary differential equations]{ODE}{ODE}{ordinary differential equation}
\newacronym[category=special,firstplural=residual networks]{RESNET}{ResNet}{residual network}
\newacronym[category=general]{SGD}{SGD}{stochastic gradient descent}
\newacronym[category=general]{SG}{SG}{stochastic gradient}
\newacronym[category=general]{ARD}{ARD}{automatic relevance determination}
\newacronym[category=general]{NNGP}{NNGP}{neural network Gaussian process}
\newacronym[category=general]{SGLD}{SGLD}{stochastic gradient Langevin dynamics}
\newacronym[category=general]{BBB}{BBB}{Bayes by Backprop}
\newacronym[category=general]{PBP}{PBP}{probabilistic backpropagation procedure}
\newacronym[category=general]{FGE}{FGE}{gast geometric ensembling}
\newacronym[category=general]{SWA}{SWA}{stochastic weight averaging}
\newacronym[category=general]{SWAG}{SWAG}{stochastic weight averaging Gaussian}
\newacronym[category=special]{PREDCP}{PREDCP}{predictive complexity prior}
\newacronym[category=general]{AI}{AI}{artificial intelligence}
\newacronym[category=general]{JFT-300M}{JFT-300M}{}
\newacronym[category=general]{WMT-2014}{JFT-300M}{}
\newacronym[category=special]{Imagenet}{Imagenet}{}
\newacronym[category=special]{MNIST}{MNIST}{}
\newacronym[category=special]{CIFAR-10}{CIFAR-10}{}
\newacronym[category=special]{CIFAR-100}{CIFAR-100}{}
\newacronym[category=special]{}{}{}
\newacronym[category=special]{CONVIT}{CONVIT}{}
\newacronym[category=general]{PD}{PD}{positive dependence}
\newacronym[category=general]{SOTA}{SOTA}{state-of-the-art}
\newacronym[category=general]{MSE}{MSE}{mean-squared error}
\newacronym[category=special]{ZERO-ONE}{ZERO-ONE}{}
\newacronym[category=special]{kfac}{K-FAC}{Kronecker-factored Approximate Curvature}
\newacronym[category=special]{tpu}{TPU}{Tensor Processing Unit}
\newacronym[category=special]{sgmcmc}{SG-MCMC}{Stochastic gradient Markov chain Monte Carlo}

\usepackage{todonotes}

\newcommand{\bE}{\boldsymbol{\mathrm{E}}}

\newcommand{\bLambda}{\boldsymbol{\Lambda}}

\newcommand{\newparagraph}[1]{\vspace{.3cm}\noindent{\textbf{#1}}}

\title{\sffamily
\textbf{\ourtitle}}
\author{Julyan Arbel$^1$, Konstantinos Pitas$^1$, Mariia Vladimirova$^1$\\
$^1$\textit{Centre Inria de l'Université Grenoble Alpes}}
\date{}

\usepackage[pagestyles]{titlesec}
\newpagestyle{mystyle}
{\sethead[\thepage][][\chaptertitle]{}{}{\thepage}}
\usepackage[Bjornstrup]{fncychap}

\begin{document}

\phantom{a}\vspace{2cm}

\noindent{\LARGE\sffamily
\textbf{A Primer on Bayesian Neural Networks:\\ \medskip
Review and Debates}}\\

\noindent{Julyan Arbel$^1$, Konstantinos Pitas$^1$, Mariia Vladimirova$^2$, Vincent Fortuin$^3$}\\

\noindent{$^1$\textit{Centre Inria de l'Université Grenoble Alpes, France}}\\
\noindent{$^2$\textit{Criteo AI Lab, Paris, France}}\\
\noindent{$^3$\textit{Helmholtz AI, Munich, Gremany}}\\
\\

\thispagestyle{empty}
\begin{center}
    \rule{16cm}{.5pt}

\begin{quote}

    Neural networks have achieved remarkable performance across various problem domains, but their widespread applicability is hindered by inherent limitations such as overconfidence in predictions, lack of interpretability, and vulnerability to adversarial attacks. To address these challenges, Bayesian neural networks (BNNs) have emerged as a compelling extension of conventional neural networks, integrating uncertainty estimation into their predictive capabilities.

This comprehensive primer presents a systematic introduction to the fundamental concepts of neural networks and Bayesian inference, elucidating their synergistic integration for the development of BNNs. The target audience comprises statisticians with a potential background in Bayesian methods but lacking deep learning expertise, as well as machine learners proficient in deep neural networks but with limited exposure to Bayesian statistics. We provide an overview of commonly employed priors, examining their impact on model behavior and performance. Additionally, we delve into the practical considerations associated with training and inference in BNNs.

Furthermore, we explore advanced topics within the realm of BNN research, acknowledging the existence of ongoing debates and controversies. By offering insights into cutting-edge developments, this primer not only equips researchers and practitioners with a solid foundation in BNNs, but also illuminates the potential applications of this dynamic field. As a valuable resource, it fosters an understanding of BNNs and their promising prospects, facilitating further advancements in the pursuit of knowledge and innovation.
\end{quote}
    \rule{16cm}{.5pt}
\end{center}

\tableofcontents

\newpage



\section{Introduction}

\newparagraph{Motivation.} 
Technological advancements have sparked an increased interest in the development of models capable of acquiring knowledge and performing tasks that resemble human abilities. These include tasks such as object recognition and scene segmentation in images, speech recognition in audio signals, and natural language understanding. They are commonly referred to as artificial intelligence (AI) tasks. AI systems possess the remarkable ability to mimic human thinking and behavior.

Machine learning, a subset of artificial intelligence, encompasses a fundamental aspect of AI—learning the underlying relationships within data and making decisions without explicit instructions. Machine learning algorithms autonomously learn and enhance their performance by leveraging their output. These algorithms do not rely on explicit instructions to generate desired outcomes; instead, they learn by analyzing accessible datasets and comparing them with examples of the desired output.

Deep learning, a specialized field within machine learning, focuses on algorithms inspired by the structure and functioning of the human brain, known as (artificial) neural networks. Deep learning concepts enable machines to acquire human-like skills. Through deep learning, computer models can be trained to perform classification tasks using inputs such as images, text, or sound. Deep learning has gained popularity due to its ability to achieve state-of-the-art performance. The training of these models involves utilizing large labeled datasets in conjunction with neural network architectures.

Neural networks, or NNs, are particularly effective deep learning models that can solve a wide range of problems. They are now widely employed across various domains. For instance, they can facilitate translation between languages, guide users in banking applications, or even generate artwork in the style of famous artists based on simple photographs. However, neural networks are often regarded as black boxes due to the lack of intuitive interpretations that would allow us to trace the flow of information from input to output.

In certain industries, the acceptance of AI algorithms necessitates explanations. This requirement may stem from regulations encompassed in the concept of AI safety or from human factors. In the field of medical diagnosis and treatment, decisions based on AI algorithms can have life-changing consequences. While AI algorithms excel at detecting various health conditions by identifying minute details imperceptible to the human eye, doctors may hesitate to rely on this technology if they cannot explain the rationale behind its outcomes.

In the realm of finance, AI algorithms can assist in tasks such as assigning credit scores, evaluating insurance claims, and optimizing investment portfolios, among other applications. However, if these algorithms produce biased outputs, it can cause reputational damage and even legal implications. Consequently, there is a pressing need for interpretability, robustness, and uncertainty estimation in AI systems.

The exceptional performance of deep learning models has fueled research efforts aimed at comprehending the mechanisms that drive their effectiveness. Nevertheless, these models remain highly opaque, as they lack the ability to provide human-understandable accounts of their reasoning processes or explanations. Understanding neural networks can significantly contribute to the development of safe and explainable AI algorithms that could be widely deployed to improve people's lives. The Bayesian perspective is often viewed as a pathway toward trustworthy AI. It employs probabilistic theory and approximation methods to express and quantify uncertainties inherent in the models. However, the practical implementation of Bayesian approaches for uncertainty quantification in deep learning models often incurs significant computational costs and necessitates the use of improved approximation techniques.

\newparagraph{Objectives and outline.}
The recent surge of research interest in Bayesian deep learning has spawned several notable review articles that contribute valuable insights to the field. For instance, \cite{jospin2020hands} present a useful contribution by offering practical implementations in Python, enhancing the accessibility of Bayesian deep learning methodologies.
Another significant review by \cite{abdar2021review} provides a comprehensive assessment of uncertainty quantification techniques in deep learning, encompassing both frequentist and Bayesian approaches. This thorough examination serves as an essential resource for researchers seeking to grasp the breadth of available methods.
While existing literature delves into various aspects of Bayesian neural networks, \cite{goan2020bayesian} specifically focuses on inference algorithms within BNNs. However, the comprehensive coverage of prior modeling, a critical component of BNNs, is not addressed in this review.
Conversely, \cite{fortuin2021priors} presents a meticulous examination of priors utilized in diverse Bayesian deep learning models, encompassing BNNs, deep Gaussian processes, and variational auto-encoders (VAEs). This review offers valuable insights into the selection and impact of priors across different Bayesian modeling paradigms.

In contrast to these works, our objective is to offer an accessible and comprehensive guide to Bayesian neural networks, catering to both statisticians and machine learning practitioners. The target audience comprises statisticians with a potential background in Bayesian methods but lacking deep learning expertise, as well as machine learners proficient in deep neural networks but with limited exposure to Bayesian statistics.
Assuming no prior familiarity with either deep learning or Bayesian statistics, we provide succinct explanations of both domains in Section~\ref{section:nn} and Section~\ref{section:bml}, respectively. These sections serve as concise reminders, enabling readers to grasp the foundations of each field.
Subsequently, in Section~\ref{section:bnn}, we delve into Bayesian neural networks, elucidating their core concepts, with a specific emphasis on frequently employed priors and inference techniques. By addressing these fundamental aspects, we equip the reader with a solid understanding of BNNs and their associated methodologies.
Furthermore, in Section~\ref{section:bayesian_and_non-bayesian}, we analyze the principal challenges encountered by contemporary Bayesian neural networks. This exploration provides readers with a comprehensive overview of the obstacles inherent to this field, highlighting areas for further investigation and improvement.
Ultimately, Section~\ref{section:conlusion} concludes our guide, summarizing the key points and emphasizing the significance of Bayesian neural networks. By offering this cohesive resource, our goal is to empower statisticians and machine learners alike, fostering a deeper understanding of BNNs and facilitating their broader application in practice.\footnote{We provide an up-to-date reading list of research articles related to Bayesian neural networks at this link: \href{https://github.com/konstantinos-p/Bayesian-Neural-Networks-Reading-List}{https://github.com/konstantinos-p/Bayesian-Neural-Networks-Reading-List}.}

\newpage 

\section{Neural networks and statistical learning theory}
\label{section:nn}

The inception of neural network models can be traced back to 1955 when the first model, known as the \textit{perceptron}, was constructed \citep{rosenblatt1958perceptron}. Subsequently, significant advancements have taken place in this field, notably the discovery of the \textit{backpropagation} algorithm in the 1980s~\citep{rumelhart1986learning}. This algorithm revolutionized neural networks by enabling efficient training through gradient-descent-based methods. 
However, the current era of profound progress in deep learning commenced in 2012 with a notable milestone: convolutional neural networks, when trained on graphics processing units (GPUs) for the first time, achieved exceptional performance on the ImageNet task~\citep{Krizhevsky2012}. This breakthrough marked a significant turning point and propelled the rapid advancement of deep learning methodologies.

\newparagraph{Definition and notations.}
\textit{Neural networks} are hierarchical models made of layers: an input, several hidden layers, and an output, see  Figure~\ref{figure:nn_visualization_intro}. The number of hidden layers $L$ is called \textit{depth}. Each layer following the input layer consists of units which are linear combinations of previous layer units transformed by a nonlinear function, often referred to as the nonlinearity or \textit{activation function} denoted by $\phi: \mathbb{R} \to \mathbb{R}$. Given an input $\bx \in \mathbb{R}^N$ (for instance an image made of $N$ pixels), the $\ell$-th hidden layer consists of two vectors whose size is called the \textit{width} of layer, denoted by $H_\ell$, where $\ell = 1, \dots, L$. The vector of units before application of the non-linearity is called  \textit{pre-nonlinearity} (or \textit{pre-activation}), and is denoted by $\bg^{(\ell)}=\bg^{(\ell)}(\bx)$, while the vector obtained after element-wise application of $\phi$  is called \textit{post-nonlinearity} (or \textit{post-activation}) and is denoted by  $\bh^{(\ell)}=\bh^{(\ell)}(\bx)$. More specifically, these vectors  are defined as
\begin{align}\label{eq:propagation}
      \bg^{(\ell)}(\bx) = \bw^{(\ell)} \bh^{(\ell - 1)} (\bx), \quad \bh^{(\ell)} (\bx) = \phi(\bg^{(\ell)}(\bx)),
\end{align}
where $\bw^{(\ell)}$ is a weight matrix of dimension $H_\ell\times H_{\ell-1}$ including a bias vector, with the convention that $H_0=N$, the input dimension. 

\tikzset{%
  input/.style={
      circle,
      draw,
      color={rgb, 255:red, 107; green, 65; blue, 144 },
      draw opacity=1,
      fill={rgb, 255:red, 107; green, 65; blue, 144 },
      fill opacity=0.48,
      minimum size=0.5cm
    },
    every neuron/.style={
      circle,
      draw,
      color={rgb, 255:red, 65; green, 117; blue, 5 },
      draw opacity=1,
      fill={rgb, 255:red, 65; green, 117; blue, 5 },
      fill opacity=0.48,
      minimum size=0.5cm
    },
    neuron missing/.style={
      draw=none, 
      fill=none,
      scale=1,
      text height=0.333cm,
      execute at begin node=\color{black}$\vdots$
    },
    output/.style={
      circle,
      draw,
      color={rgb, 255:red, 107; green, 65; blue, 144 },
      draw opacity=1,
      fill={rgb, 255:red, 107; green, 65; blue, 144 },
      fill opacity=0.48,
      minimum size=0.5cm
    },
}
\begin{figure*}[ht!]
\begin{center}
\scalebox{.85}{
\begin{tikzpicture}[x=1.2cm, y=0.8cm, >=stealth]

\foreach \m/\l [count=\y] in {1,2,3,missing,4}
  \node [input/.try, neuron \m/.try] (input-\m) at (0,2.2-\y) {};

\foreach \m [count=\y] in {1,2,missing,3}
  \node [every neuron/.try, neuron \m/.try ] (firsthidden-\m) at (2,1.9-\y*1.1) {};

\foreach \m [count=\y] in {1,2,missing,3}
  \node [every neuron/.try, neuron \m/.try ] (secondhidden-\m) at (4,1.9-\y*1.1) {};

\foreach \m [count=\y] in {1,2,3,4}
  \node [neuron missing/.try, neuron \m/.try ] (thirdhidden-\m) at (6,1.9-\y*1.1) {};

\foreach \m [count=\y] in {1,2,missing,3}
  \node [every neuron/.try, neuron \m/.try ] 
(lasthidden-\m) at (8,1.9-\y*1.1) {};

\foreach \m [count=\y] in {1,missing,2}
  \node [output/.try, neuron \m/.try ] 
(output-\m) at (10,1.6-\y*1.1) {};

\foreach \l [count=\i] in {1,2,3,n}
  \draw [<-] (input-\i) -- ++(-1,0);
  \node [above] at (input-1.north) {$\bx$};

\foreach \l [count=\i] in {1};
\node [above] at (firsthidden-1.north) {$\bh^{(1)}$};


\foreach \l [count=\i] in {1};
  \node [above] at (secondhidden-1.north) {$\bh^{(2)}$};

\foreach \l [count=\i] in {1}
  \node [above] at (thirdhidden-1.north) {$\bh^{(\ell)}$};

\foreach \l [count=\i] in {1,2,n}
 \node [above] at (lasthidden-1.north) {$\bh^{(L-1)}$};

\foreach \l [count=\i] in {1}
  \node [above] at (output-1.north) {$\by$};

\foreach \i in {1,...,4}
  \foreach \j in {1,...,3}
    \draw [->] (input-\i) -- (firsthidden-\j);

\foreach \i in {1,...,3}
  \foreach \j in {1,...,3}
    \draw [->] (firsthidden-\i) -- (secondhidden-\j);

\foreach \i in {1,...,3}
  \foreach \j in {1,...,4}
    \draw [->] (secondhidden-\i) -- (thirdhidden-\j);

\foreach \i in {1,...,4}
  \foreach \j in {1,...,3}
    \draw [->] (thirdhidden-\i) -- (lasthidden-\j);

\foreach \i in {1,...,3}
  \foreach \j in {1,...,2}
    \draw [->] (lasthidden-\i) -- (output-\j);

\foreach \l [count=\x from 0] in {\footnotesize{input}, \footnotesize{$1^{\text{st}}$ hidden}, \footnotesize{$2^{\text{nd}}$ hidden}, , \footnotesize{last hidden}, \footnotesize{output}}
  \node [align=center, above] at (\x*2,2) {\l};
\end{tikzpicture}
}
\end{center}
\caption{Simple fully-connected neural network architecture.}
\label{figure:nn_visualization_intro}
\end{figure*}
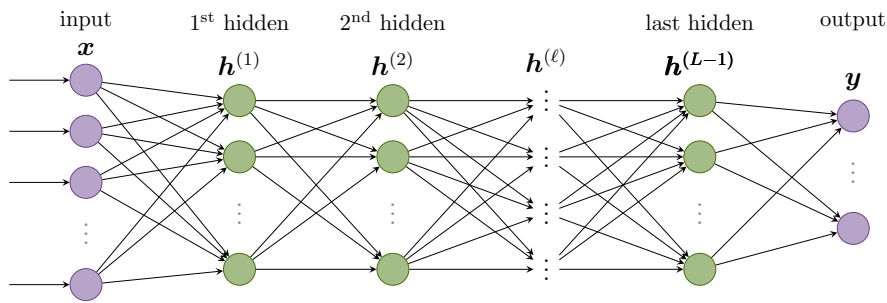

\newparagraph{Supervised learning.} We denote the learning sample $(X,Y)=\{(\bx_i,\by_i)\}^n_{i=1}\in(\mathcal{X}\times\mathcal{Y})^n$, which contains $n$ input-output pairs. Observations $(X,Y)$ are assumed to be randomly sampled from a distribution $\mathfrak{D}$. Thus, we denote $(X,Y)\sim\mathfrak{D}^n$ the i.i.d observation of $n$ elements. We define the test set $(X_{\mathrm{test}},Y_{\mathrm{test}})$ of $n_{\mathrm{test}}$ samples in a similar way to that of the learning sample. We consider some loss function $\mathcal{L}:\mathcal{F}\times\mathcal{X}\times\mathcal{Y}\rightarrow\mathbb{R}$, where $\mathcal{F}$ is a set of predictors $f:\mathcal{X}\rightarrow\mathcal{Y}$. 
We also denote the empirical risk $\mathcal{R}^{\mathcal{L}}_n(f)=(1/n)\sum_i \mathcal{L}(f(\bx_i), \by_i)$ and the risk 
\begin{equation}
\label{eq:risk}
    \mathcal{R}^{\mathcal{L}}_{\mathfrak{D}}(f)=\bE_{(\bx,\by)\sim\mathfrak{D}}\,\mathcal{L}(f(\bx), \by).
\end{equation}
The minimizer of $\mathcal{R}^{\mathcal{L}}_{\mathfrak{D}}$ is called \textit{Bayes optimal predictor} $f^* = \argmin_{f : \mathcal{X} \to \mathcal{Y}} \mathcal{R}^{\mathcal{L}}_{\mathfrak{D}}(f)$. The minimal risk $\mathcal{R}^{\mathcal{L}}(f^*)$,  called \textit{Bayes risk}, is achieved by the Bayes optimal predictor $f^*$.

Returning to neural networks, we denote their vectorized weights by $\bw\in\mathbb{R}^{d}$ with $d = \sum_{\ell = 1}^L H_{\ell - 1} H_\ell$, such that $f(\bx) = f_{\bw}(\bx)$. The goal is to find the optimal weights such that the neural network output $\by_i^*$ for input $\bx_i$ is the \textit{closest} to the given label $\by_i$, which is estimated by a loss function $\mathcal{L}$. In the regression problem, for example, the loss function $\mathcal{L}$ could be the mean-squared error $\|\by_i^* - \by_i\|^2$. 
The optimization problem is then to minimize the empirical risk:
\begin{equation*}
    \hat \bw = \argmin_{\bw} \mathcal{R}^{\mathcal{L}}_n(f_{\bw}).
\end{equation*}
With optimal weights $\hat \bw$, the empirical risk $\mathcal{R}^{\mathcal{L}}_{n}(\hat f)$ is small and should be close to the Bayes risk $\mathcal{R}^{\mathcal{L}}(f^*)$.

\newparagraph{Training.}
The main workhorse of neural network training is gradient-based optimization: 
\begin{equation}
    \label{eq:sgd}
    \bw \leftarrow \bw - \eta\, \nabla_{\bw} \mathcal{R}^{\mathcal{L}}_n(f_{\bw}),
\end{equation}
where $\eta > 0$ is a \textit{step size}, or \textit{learning rate}, and the gradients are computed as products of gradients between each layer \textit{from right to left}, a procedure called \textit{backpropagation}~\citep{rumelhart1986learning}, thus making use of the chain rule and efficient implementations for matrix-vector products. For large datasets, this optimization is often replaced by \gls{SGD}, where gradients are approximated on some randomly chosen subsets called \textit{batches}~\citep{robbins1951stochastic}. In this case, it requires a careful choice of the learning rate parameter. For a survey on different optimization methods, see, for example, \cite{sun2019survey}. For the optimization procedure, another important aspect is how to choose the weight initialization; we discuss this in detail in Section~\ref{section:initialization}. 

\subsection{Choice of architecture} 

With the progress in deep learning,  different neural network architectures have been introduced to better adapt to different learning problems. Knowledge about the data allows encoding specific properties into the architecture. Depending on the architecture, this results (among other benefits) in better feature extraction, a reduced number of parameters, invariance or equivariance to certain transformations, robustness to distribution shifts and more numerically stable optimization procedures. We shortly review some important models and refer the reader to~\cite{sarker2021deep} for a more in-depth overview of recent techniques.

\textit{\glsunset{CNN}Convolutional neural networks}
(\glspl{CNN}) are widely used in computer vision. Image data has spatial features that refer to the arrangement of pixels and their relationship. For example, we can easily identify a human’s face by looking at specific features like eyes, nose, mouth, etc. 
\glspl{CNN} were introduced to capture spatial features by using \textit{convolutional layers}, a particular case of the fully-connected layers described above, where certain sets of parameters are shared~\citep{lecun1989backpropagation,Krizhevsky2012}. Convolutional layers perform a dot product of a convolution kernel with the layer's input matrix. As the convolution kernel slides along the input matrix for the layer, the convolution operation generates a feature map of smaller dimension which serves as an input to the next layer. It introduces the concept of parameter sharing where the same kernel, or filter, is applied across different input parts to extract the relevant features from the input. 

\textit{\glsunset{RNN}Recurrent neural networks} (\glspl{RNN}) are designed to save the output of a layer by adding it back to the input~\citep{rumelhart1986learning,hochreiter1997long}. During training, the recurrent layer has some information from the previous time-step. Such neural networks are advantageous for sequential data where each sample can be assumed to be dependent on preceding ones. 

\textit{\glsunset{RESNET}Residual neural networks} (\glspl{RESNET}) have residual blocks which add the output from the previous layer to the output of the current layer, a so-called \textit{skip-connection}~\citep{he2016deep}. It allows training very deep neural networks by ensuring that deeper layers in the model will perform at least as well as layers preceding them.~\citep{he2016identity}.

\textit{Transformers} are a type of neural network architecture that is almost entirely based on the \textit{attention mechanism}~\citep{vaswani2017attention}. The idea behind \textit{attention} is to find and focus on small, but important, parts of the input data. Transformers show better results than convolutional or residual networks on some tasks with big datasets such as image classification with JFT-300M (300M images), or English-French machine translation with WMT-2014 (36M sentences, split into a 32000 token vocabulary).

An open question in deep learning is why deep \glspl{NN} achieve state-of-the-art performance in a significant number of applications. The common belief is that neural networks' complexity and over-parametrization result in tremendous \textit{expressive power}, beneficial \textit{inductive bias}, flexibility to avoid \textit{overfitting} and, therefore, the ability to \textit{generalize} well. Yet, the high dimensionalities of the data and parameter spaces of these models make them challenging to understand theoretically. In the following, we review these open topics of research as well as the current scientific consensus on them.

\subsection{Expressiveness}

The expressive power describes neural networks’ ability to approximate functions. 
In the late 1980s, a line of works established a universal approximation theorem, stating that one-hidden-layer neural networks with a suitable activation function could approximate any continuous function on a compact domain, that is $f: [0,1]^N \to \mathbb{R}$, to any desired accuracy~\citep{cybenko1989approximation, funahashi1989approximate, hornik1989multilayer,barron1994approximation}. The obstacle is that the size of such networks may be exponential in the input dimension $N$, which makes them highly prone to overfitting as well as impractical, since adding extra layers in the model is often a cheaper way to increase the representational power of the neural network. 
More recently, \cite{telgarsky2016benefits} studied which functions neural networks could represent by focusing on the choice of the architecture and showed that deeper models are more expressive. \cite{chatziafratis2020depth,chatziafratis2020better} extended this result by obtaining width-depth trade-offs. 

Another approach is to analyze the finite-sample expressiveness of neural networks. \cite{zhang2017understanding} state that as soon as the number of parameters of a network is greater than the input sample size, even a simple two-layer neural network can represent any function of the input sample. Though neural networks are theoretically expressive, the core of the learning problem lies in their complexity, and research focuses on obtaining complexity bounds.   


In general, the ability to approximate or to \textit{express} specific functions can be considered as explicit \textit{inductive bias} which we discuss in detail in the next section. 

\subsection{Inductive bias}
\label{section:inductive_bias}
By choosing a design and a training procedure for a model assigned to a given problem, we make some assumptions on the problem structure. These assumptions are summed in the term \textit{inductive bias}\footnote{ The term \textit{inductive} comes from philosophy:  \textit{inductive reasoning} refers to \textit{generalization} from specific observations to a conclusion. This is a counterpoint to \textit{deductive reasoning}, which refers to \textit{specialization} from general ideas to a conclusion.}, i.e., prior preferences for specific models and problems.

\begin{figure*}[ht!]
\vspace*{2ex}
\hspace*{2ex}
linear regression 
\hspace*{24ex}
$k$-nearest neighbours\\
\vspace*{1ex}
    \centering
    \includegraphics[width=0.45\textwidth]{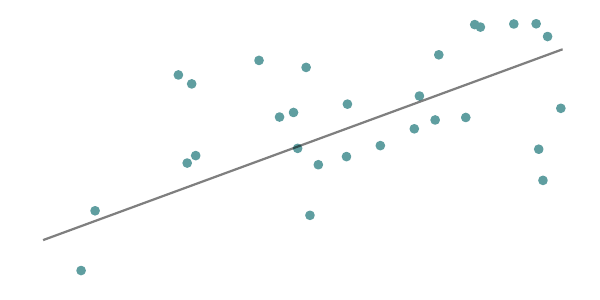}
    \includegraphics[width=0.45\textwidth]{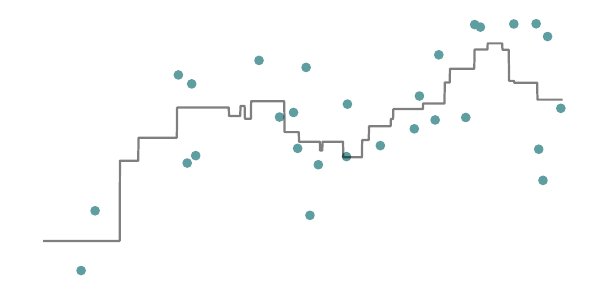}
    \caption{Example of using the linear regression (left) and $k$-nearest neighbours regression (right) models on simulated data points.}
    \label{fig:inductive_bias_examples}
\end{figure*}

\newparagraph{Examples.} For instance, the linear regression model is built on the assumption of a linear relationship between the target variable and the features. The knowledge that the data is of a linear nature is \textit{embedded} into the model. Because of this limitation of the linearity of the model, linear regression is bound to perform poorly for data in where the target variable does not linearly depend on features, see the left plot of Figure~\ref{fig:inductive_bias_examples}. This assumption of a linear relationship between the target and the features is the inductive bias of linear regression. In the $k$-nearest neighbours model, the inductive bias is that the answer for any object should be calculated only on the basis of what values of the answers were in the elements of the training sample closest to this object, see the right plot of Figure~\ref{fig:inductive_bias_examples}. In the non-linear regression, the assumption is some non-linear function. 

\newparagraph{Importance.} 
The goal of a machine learning model is to derive a general rule for all elements of a domain based on a limited number of observations. In other words, we want the model to \textit{generalize} to data it has not seen before. Such generalization is impossible without the presence of \textit{inductive bias} in the model because the training sample is always \textit{finite}. From a finite set of observations, without making any additional assumptions about the data, a general rule can be deduced in an infinite number of ways. Inductive bias is additional information about the nature of the data for the model; a way to show models \textit{which way to think}. It allows the model to prioritize one generalization method over another. Thus, when choosing a model for training to solve a specific problem, one needs to choose a model whose inductive bias better matches the nature of the data and better allows to solve this problem. The introduction of any inductive bias into a machine learning model relies on certain characteristics of the model \textit{architecture}, \textit{training algorithm} and manipulations on \textit{training data}.


\newparagraph{Inductive bias and training data.} One can also consider inductive bias through training data. 
The fewer data, the more likely the model chooses a poor generalization method. If the training sample is small, models such as neural networks are often \textit{overfitted}. For example, when solving the problem of classifying images of cats and dogs, sometimes attention is paid to the background and not to the animals themselves. But people, unlike neural networks, can quickly learn on the problem of classifying cats and dogs, having only a dozen pictures in the training set. This is because people have additional inductive bias: we know that there is a background in the picture, and there is an object, and during the classification of pictures, you need to pay attention only to the object itself. And the neural network before training does not know about any “backgrounds” and “objects”---it is simply given different pictures and asked to learn how to distinguish them. Thus, \textit{the smaller the training sample and the more complex the problem, the stronger inductive bias} is required to be invested in the model device for successful model training.

Conversely, the more extensive and more diverse the training set, the more knowledge about the nature of the data the model receives during training. This means that the less likely the model is to choose a “bad” generalization method that will work poorly on data outside the training set. 
Thus, \textit{the more data you have, the better the model will train}. 

One of the tricks to increase the dataset is to artificially augment the training set by introducing distortions into the inputs, a procedure known as \textit{data augmentation}. Suppose we are trying to classify images of objects or handwritten digits. Each time we visit a training example, we can randomly distort it, for instance, by shifting it by a few pixels, adding noise, rotating it slightly, or applying some sort of warping. 
This can increase the effective size of the training set and make it more likely that any given test example has a closely related training example. The data augmentation procedure is a sort of inductive bias because it requires the knowledge of how to construct additional data points, such as if the object or part of the object can be rotated, zoomed in, etc.  

\newparagraph{Inductive bias and simplicity.} The \textit{no free lunch} theorem states that no single learning algorithm can succeed on all possible problems~\citep{wolpert1996lack}. It is, thus, essential to enforce a form of \textit{simplicity} in the algorithm, typically by restricting the class of models to be learned, which may reflect prior knowledge about the problem being tackled. This is associated with \textit{inductive bias} which should encode the prior knowledge to seek for efficiency. In the context of neural networks, one form of simplicity is in the choice of \textit{architecture}, such as using convolutional neural networks~\citep{lecun1989backpropagation} when learning from image data. Another example is \textit{sparsity}, which may seek models that only rely on a few relevant variables out of many available ones and can be achieved through some regularization methods~\citep{tibshirani1996regression}.

\newparagraph{Inductive bias of neural network architecture.} 
A number of deep neural network architectures have been designed with the aim of improving the inductive bias of the corresponding predictor. Here we review two popular neural network architectures that encode useful inductive biases.

\textit{\glsunset{CNN}Convolutional neural networks} (\glspl{CNN}). The inductive bias of convolutional layers~\citep{lecun1989backpropagation} is the assumption of compactness and translation invariance. The convolution filter is designed in such a way that at one time it captures a compact part of the entire image (for example, a $3\times3$ pixels square), regardless of the distant pixels of the image. Also in the convolutional layer, the same filter is used to process the entire image (the same filter processes all $3\times3$ pixels square). It turns out that the convolutional layer is designed in such a way that its inductive bias correlates well with the nature of images and objects on them, which is why convolutional neural networks are so efficient at processing images~\citep{Krizhevsky2012}. This is an example of the desired, or \textit{explicit} inductive bias. \textit{What makes data efficiently learnable by fitting a huge neural network with a specific algorithm? Is there implicit inductive bias?} \cite{ulyanov2018deep} demonstrate that the output of a convolutional neural network with randomly initialized weights corresponds to a \textit{deep image prior}, i.e. non-trivial image properties, \textit{before} training. It means that how convolutional neural networks are designed, their architecture itself, helps to encode the information from images. \cite{geirhos2019imagenet} show that \textit{convolutional neural networks} have implicit inductive bias concerning the texture of images: it turns out that convolutional networks are designed in such a way that when processing images, they pay more attention to textures rather than to the shapes of objects. To get rid of this undesirable behavior, the images from the training dataset are augmented so that the dataset contains more images of the same shape, but with different types of textures~\citep{li2021shape}. Despite the popularity of the topic, the implicit inductive bias in neural networks is still an open question due to the complexity of the models. 

\textit{Visual transformers}~\citep{dosovitskiy2021image} are a type of neural network architecture that shows better results than convolutional networks on some tasks, including, for example, classification of images from the \acrshort{JFT-300M} dataset. This dataset consists of 300 million images, while \acrshort{Imagenet} has 1.2 million images. The visual transformer is almost entirely based on the \textit{attention mechanism}~\citep{vaswani2017attention}, so the model has the inductive bias that attention has which consists in a shift towards simpler functions. But like convolutions, transformers also have the implicit inductive bias of neural networks~\citep{morrison2021exploring}. Though there is still a lot of ongoing research on transformers, the inductive bias of transformers is much simpler than that of convolutional neural networks, as the former models impose fewer restrictions than the latter models. Here we see confirmation that the larger dataset we have at our disposal, the less inductive bias is required, and the better the model can learn for the task. Therefore, transformers have simple inductive bias and show state-of-the-art results in image processing, but they require a lot of data. On the contrary, convolutional neural networks have a strong inductive bias, and they perform well on smaller datasets. Recently, \cite{d2021convit} combined the transformer and convolutional neural network architectures, introducing the \acrshort{CONVIT} model. This model is able to process images almost as well as transformers, while requiring less training data.


\subsection{Generalization and overfitting}
\label{section:generalization_and_overfitting}

When we train a machine learning model, we do not just want it to learn to model the training data. We want it to \textit{generalize} to data it has not seen before. Fortunately, there is a way to measure an algorithm’s generalization performance: we measure its performance on a held-out test set, consisting of examples it has not seen before. If an algorithm works well on the training set but fails to generalize, we say it suffers from \textit{overfitting}. Modern machine learning systems based on deep neural networks are usually over-parameterized, i.e., the number of parameters in the model is much larger than the size of the training data, which makes these systems prone to overfitting.

\newparagraph{Classical regime.} Let us randomly divide the original dataset into a train, validation and test set. The model is trained by optimizing the training error computed on the train set, then its performance is checked by computing the validation error on the validation set. After tuning any existing hyperparameters by checking the validation error, the model (or models) are then evaluated on final time on the test set. 

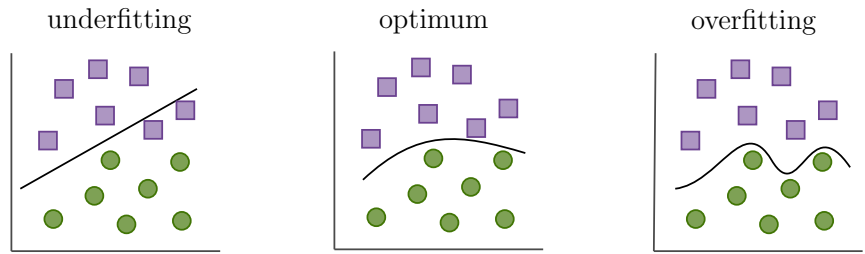
\begin{figure*}[ht!]
\centering 
\tikzset{every picture/.style={line width=0.75pt}} 
\scalebox{.9}{
\begin{tikzpicture}[x=0.75pt,y=0.75pt,yscale=-1,xscale=1]

\draw [color={rgb, 255:red, 0; green, 0; blue, 0 }  ,draw opacity=1 ]   (35.5,125) -- (134.5,69) ;
\draw    (402.5,125) .. controls (427,122) and (440.5,82) .. (456.5,109) .. controls (472.5,136) and (477.5,79) .. (500.5,113) ;
\draw    (227.5,120) .. controls (258.5,91) and (281.5,94) .. (318.5,105) ;
\draw [color={rgb, 255:red, 74; green, 74; blue, 74 }  ,draw opacity=1 ]   (30.5,160) -- (147.5,160) ;
\draw [color={rgb, 255:red, 74; green, 74; blue, 74 }  ,draw opacity=1 ]   (30.5,160) -- (30.5,49) ;
\draw  [color={rgb, 255:red, 65; green, 117; blue, 5 }  ,draw opacity=1 ][fill={rgb, 255:red, 65; green, 117; blue, 5 }  ,fill opacity=0.68 ] (49,142) .. controls (49,139.24) and (51.24,137) .. (54,137) .. controls (56.76,137) and (59,139.24) .. (59,142) .. controls (59,144.76) and (56.76,147) .. (54,147) .. controls (51.24,147) and (49,144.76) .. (49,142) -- cycle ;
\draw  [color={rgb, 255:red, 65; green, 117; blue, 5 }  ,draw opacity=1 ][fill={rgb, 255:red, 65; green, 117; blue, 5 }  ,fill opacity=0.68 ] (72,129) .. controls (72,126.24) and (74.24,124) .. (77,124) .. controls (79.76,124) and (82,126.24) .. (82,129) .. controls (82,131.76) and (79.76,134) .. (77,134) .. controls (74.24,134) and (72,131.76) .. (72,129) -- cycle ;
\draw  [color={rgb, 255:red, 65; green, 117; blue, 5 }  ,draw opacity=1 ][fill={rgb, 255:red, 65; green, 117; blue, 5 }  ,fill opacity=0.68 ] (81,109) .. controls (81,106.24) and (83.24,104) .. (86,104) .. controls (88.76,104) and (91,106.24) .. (91,109) .. controls (91,111.76) and (88.76,114) .. (86,114) .. controls (83.24,114) and (81,111.76) .. (81,109) -- cycle ;
\draw  [color={rgb, 255:red, 65; green, 117; blue, 5 }  ,draw opacity=1 ][fill={rgb, 255:red, 65; green, 117; blue, 5 }  ,fill opacity=0.68 ] (90,145) .. controls (90,142.24) and (92.24,140) .. (95,140) .. controls (97.76,140) and (100,142.24) .. (100,145) .. controls (100,147.76) and (97.76,150) .. (95,150) .. controls (92.24,150) and (90,147.76) .. (90,145) -- cycle ;
\draw  [color={rgb, 255:red, 65; green, 117; blue, 5 }  ,draw opacity=1 ][fill={rgb, 255:red, 65; green, 117; blue, 5 }  ,fill opacity=0.68 ] (102,125) .. controls (102,122.24) and (104.24,120) .. (107,120) .. controls (109.76,120) and (112,122.24) .. (112,125) .. controls (112,127.76) and (109.76,130) .. (107,130) .. controls (104.24,130) and (102,127.76) .. (102,125) -- cycle ;
\draw  [color={rgb, 255:red, 65; green, 117; blue, 5 }  ,draw opacity=1 ][fill={rgb, 255:red, 65; green, 117; blue, 5 }  ,fill opacity=0.68 ] (120,110) .. controls (120,107.24) and (122.24,105) .. (125,105) .. controls (127.76,105) and (130,107.24) .. (130,110) .. controls (130,112.76) and (127.76,115) .. (125,115) .. controls (122.24,115) and (120,112.76) .. (120,110) -- cycle ;
\draw  [color={rgb, 255:red, 65; green, 117; blue, 5 }  ,draw opacity=1 ][fill={rgb, 255:red, 65; green, 117; blue, 5 }  ,fill opacity=0.68 ] (121,143) .. controls (121,140.24) and (123.24,138) .. (126,138) .. controls (128.76,138) and (131,140.24) .. (131,143) .. controls (131,145.76) and (128.76,148) .. (126,148) .. controls (123.24,148) and (121,145.76) .. (121,143) -- cycle ;
\draw  [color={rgb, 255:red, 107; green, 65; blue, 144 }  ,draw opacity=1 ][fill={rgb, 255:red, 107; green, 65; blue, 144 }  ,fill opacity=0.55 ] (46,93) -- (56,93) -- (56,103) -- (46,103) -- cycle ;
\draw  [color={rgb, 255:red, 107; green, 65; blue, 144 }  ,draw opacity=1 ][fill={rgb, 255:red, 107; green, 65; blue, 144 }  ,fill opacity=0.55 ] (55,64) -- (65,64) -- (65,74) -- (55,74) -- cycle ;
\draw  [color={rgb, 255:red, 107; green, 65; blue, 144 }  ,draw opacity=1 ][fill={rgb, 255:red, 107; green, 65; blue, 144 }  ,fill opacity=0.55 ] (78,79) -- (88,79) -- (88,89) -- (78,89) -- cycle ;
\draw  [color={rgb, 255:red, 107; green, 65; blue, 144 }  ,draw opacity=1 ][fill={rgb, 255:red, 107; green, 65; blue, 144 }  ,fill opacity=0.55 ] (74,53) -- (84,53) -- (84,63) -- (74,63) -- cycle ;
\draw  [color={rgb, 255:red, 107; green, 65; blue, 144 }  ,draw opacity=1 ][fill={rgb, 255:red, 107; green, 65; blue, 144 }  ,fill opacity=0.55 ] (97,57) -- (107,57) -- (107,67) -- (97,67) -- cycle ;
\draw  [color={rgb, 255:red, 107; green, 65; blue, 144 }  ,draw opacity=1 ][fill={rgb, 255:red, 107; green, 65; blue, 144 }  ,fill opacity=0.55 ] (105,87) -- (115,87) -- (115,97) -- (105,97) -- cycle ;
\draw  [color={rgb, 255:red, 107; green, 65; blue, 144 }  ,draw opacity=1 ][fill={rgb, 255:red, 107; green, 65; blue, 144 }  ,fill opacity=0.55 ] (123,76) -- (133,76) -- (133,86) -- (123,86) -- cycle ;
\draw [color={rgb, 255:red, 74; green, 74; blue, 74 }  ,draw opacity=1 ]   (390.5,160) -- (507.5,160) ;
\draw [color={rgb, 255:red, 74; green, 74; blue, 74 }  ,draw opacity=1 ]   (390.5,160) -- (391.5,49) ;
\draw  [color={rgb, 255:red, 65; green, 117; blue, 5 }  ,draw opacity=1 ][fill={rgb, 255:red, 65; green, 117; blue, 5 }  ,fill opacity=0.68 ] (409,142) .. controls (409,139.24) and (411.24,137) .. (414,137) .. controls (416.76,137) and (419,139.24) .. (419,142) .. controls (419,144.76) and (416.76,147) .. (414,147) .. controls (411.24,147) and (409,144.76) .. (409,142) -- cycle ;
\draw  [color={rgb, 255:red, 65; green, 117; blue, 5 }  ,draw opacity=1 ][fill={rgb, 255:red, 65; green, 117; blue, 5 }  ,fill opacity=0.68 ] (432,129) .. controls (432,126.24) and (434.24,124) .. (437,124) .. controls (439.76,124) and (442,126.24) .. (442,129) .. controls (442,131.76) and (439.76,134) .. (437,134) .. controls (434.24,134) and (432,131.76) .. (432,129) -- cycle ;
\draw  [color={rgb, 255:red, 65; green, 117; blue, 5 }  ,draw opacity=1 ][fill={rgb, 255:red, 65; green, 117; blue, 5 }  ,fill opacity=0.68 ] (441,109) .. controls (441,106.24) and (443.24,104) .. (446,104) .. controls (448.76,104) and (451,106.24) .. (451,109) .. controls (451,111.76) and (448.76,114) .. (446,114) .. controls (443.24,114) and (441,111.76) .. (441,109) -- cycle ;
\draw  [color={rgb, 255:red, 65; green, 117; blue, 5 }  ,draw opacity=1 ][fill={rgb, 255:red, 65; green, 117; blue, 5 }  ,fill opacity=0.68 ] (450,145) .. controls (450,142.24) and (452.24,140) .. (455,140) .. controls (457.76,140) and (460,142.24) .. (460,145) .. controls (460,147.76) and (457.76,150) .. (455,150) .. controls (452.24,150) and (450,147.76) .. (450,145) -- cycle ;
\draw  [color={rgb, 255:red, 65; green, 117; blue, 5 }  ,draw opacity=1 ][fill={rgb, 255:red, 65; green, 117; blue, 5 }  ,fill opacity=0.68 ] (462,125) .. controls (462,122.24) and (464.24,120) .. (467,120) .. controls (469.76,120) and (472,122.24) .. (472,125) .. controls (472,127.76) and (469.76,130) .. (467,130) .. controls (464.24,130) and (462,127.76) .. (462,125) -- cycle ;
\draw  [color={rgb, 255:red, 65; green, 117; blue, 5 }  ,draw opacity=1 ][fill={rgb, 255:red, 65; green, 117; blue, 5 }  ,fill opacity=0.68 ] (480,110) .. controls (480,107.24) and (482.24,105) .. (485,105) .. controls (487.76,105) and (490,107.24) .. (490,110) .. controls (490,112.76) and (487.76,115) .. (485,115) .. controls (482.24,115) and (480,112.76) .. (480,110) -- cycle ;
\draw  [color={rgb, 255:red, 65; green, 117; blue, 5 }  ,draw opacity=1 ][fill={rgb, 255:red, 65; green, 117; blue, 5 }  ,fill opacity=0.68 ] (481,143) .. controls (481,140.24) and (483.24,138) .. (486,138) .. controls (488.76,138) and (491,140.24) .. (491,143) .. controls (491,145.76) and (488.76,148) .. (486,148) .. controls (483.24,148) and (481,145.76) .. (481,143) -- cycle ;
\draw  [color={rgb, 255:red, 107; green, 65; blue, 144 }  ,draw opacity=1 ][fill={rgb, 255:red, 107; green, 65; blue, 144 }  ,fill opacity=0.55 ] (406,93) -- (416,93) -- (416,103) -- (406,103) -- cycle ;
\draw  [color={rgb, 255:red, 107; green, 65; blue, 144 }  ,draw opacity=1 ][fill={rgb, 255:red, 107; green, 65; blue, 144 }  ,fill opacity=0.55 ] (415,64) -- (425,64) -- (425,74) -- (415,74) -- cycle ;
\draw  [color={rgb, 255:red, 107; green, 65; blue, 144 }  ,draw opacity=1 ][fill={rgb, 255:red, 107; green, 65; blue, 144 }  ,fill opacity=0.55 ] (438,79) -- (448,79) -- (448,89) -- (438,89) -- cycle ;
\draw  [color={rgb, 255:red, 107; green, 65; blue, 144 }  ,draw opacity=1 ][fill={rgb, 255:red, 107; green, 65; blue, 144 }  ,fill opacity=0.55 ] (434,53) -- (444,53) -- (444,63) -- (434,63) -- cycle ;
\draw  [color={rgb, 255:red, 107; green, 65; blue, 144 }  ,draw opacity=1 ][fill={rgb, 255:red, 107; green, 65; blue, 144 }  ,fill opacity=0.55 ] (457,57) -- (467,57) -- (467,67) -- (457,67) -- cycle ;
\draw  [color={rgb, 255:red, 107; green, 65; blue, 144 }  ,draw opacity=1 ][fill={rgb, 255:red, 107; green, 65; blue, 144 }  ,fill opacity=0.55 ] (465,87) -- (475,87) -- (475,97) -- (465,97) -- cycle ;
\draw  [color={rgb, 255:red, 107; green, 65; blue, 144 }  ,draw opacity=1 ][fill={rgb, 255:red, 107; green, 65; blue, 144 }  ,fill opacity=0.55 ] (483,76) -- (493,76) -- (493,86) -- (483,86) -- cycle ;
\draw [color={rgb, 255:red, 74; green, 74; blue, 74 }  ,draw opacity=1 ]   (211.5,159) -- (328.5,159) ;
\draw [color={rgb, 255:red, 74; green, 74; blue, 74 }  ,draw opacity=1 ]   (211.5,159) -- (211.5,48) ;
\draw  [color={rgb, 255:red, 65; green, 117; blue, 5 }  ,draw opacity=1 ][fill={rgb, 255:red, 65; green, 117; blue, 5 }  ,fill opacity=0.68 ] (230,141) .. controls (230,138.24) and (232.24,136) .. (235,136) .. controls (237.76,136) and (240,138.24) .. (240,141) .. controls (240,143.76) and (237.76,146) .. (235,146) .. controls (232.24,146) and (230,143.76) .. (230,141) -- cycle ;
\draw  [color={rgb, 255:red, 65; green, 117; blue, 5 }  ,draw opacity=1 ][fill={rgb, 255:red, 65; green, 117; blue, 5 }  ,fill opacity=0.68 ] (253,128) .. controls (253,125.24) and (255.24,123) .. (258,123) .. controls (260.76,123) and (263,125.24) .. (263,128) .. controls (263,130.76) and (260.76,133) .. (258,133) .. controls (255.24,133) and (253,130.76) .. (253,128) -- cycle ;
\draw  [color={rgb, 255:red, 65; green, 117; blue, 5 }  ,draw opacity=1 ][fill={rgb, 255:red, 65; green, 117; blue, 5 }  ,fill opacity=0.68 ] (262,108) .. controls (262,105.24) and (264.24,103) .. (267,103) .. controls (269.76,103) and (272,105.24) .. (272,108) .. controls (272,110.76) and (269.76,113) .. (267,113) .. controls (264.24,113) and (262,110.76) .. (262,108) -- cycle ;
\draw  [color={rgb, 255:red, 65; green, 117; blue, 5 }  ,draw opacity=1 ][fill={rgb, 255:red, 65; green, 117; blue, 5 }  ,fill opacity=0.68 ] (271,144) .. controls (271,141.24) and (273.24,139) .. (276,139) .. controls (278.76,139) and (281,141.24) .. (281,144) .. controls (281,146.76) and (278.76,149) .. (276,149) .. controls (273.24,149) and (271,146.76) .. (271,144) -- cycle ;
\draw  [color={rgb, 255:red, 65; green, 117; blue, 5 }  ,draw opacity=1 ][fill={rgb, 255:red, 65; green, 117; blue, 5 }  ,fill opacity=0.68 ] (283,124) .. controls (283,121.24) and (285.24,119) .. (288,119) .. controls (290.76,119) and (293,121.24) .. (293,124) .. controls (293,126.76) and (290.76,129) .. (288,129) .. controls (285.24,129) and (283,126.76) .. (283,124) -- cycle ;
\draw  [color={rgb, 255:red, 65; green, 117; blue, 5 }  ,draw opacity=1 ][fill={rgb, 255:red, 65; green, 117; blue, 5 }  ,fill opacity=0.68 ] (301,109) .. controls (301,106.24) and (303.24,104) .. (306,104) .. controls (308.76,104) and (311,106.24) .. (311,109) .. controls (311,111.76) and (308.76,114) .. (306,114) .. controls (303.24,114) and (301,111.76) .. (301,109) -- cycle ;
\draw  [color={rgb, 255:red, 65; green, 117; blue, 5 }  ,draw opacity=1 ][fill={rgb, 255:red, 65; green, 117; blue, 5 }  ,fill opacity=0.68 ] (302,142) .. controls (302,139.24) and (304.24,137) .. (307,137) .. controls (309.76,137) and (312,139.24) .. (312,142) .. controls (312,144.76) and (309.76,147) .. (307,147) .. controls (304.24,147) and (302,144.76) .. (302,142) -- cycle ;
\draw  [color={rgb, 255:red, 107; green, 65; blue, 144 }  ,draw opacity=1 ][fill={rgb, 255:red, 107; green, 65; blue, 144 }  ,fill opacity=0.55 ] (227,92) -- (237,92) -- (237,102) -- (227,102) -- cycle ;
\draw  [color={rgb, 255:red, 107; green, 65; blue, 144 }  ,draw opacity=1 ][fill={rgb, 255:red, 107; green, 65; blue, 144 }  ,fill opacity=0.55 ] (236,63) -- (246,63) -- (246,73) -- (236,73) -- cycle ;
\draw  [color={rgb, 255:red, 107; green, 65; blue, 144 }  ,draw opacity=1 ][fill={rgb, 255:red, 107; green, 65; blue, 144 }  ,fill opacity=0.55 ] (259,78) -- (269,78) -- (269,88) -- (259,88) -- cycle ;
\draw  [color={rgb, 255:red, 107; green, 65; blue, 144 }  ,draw opacity=1 ][fill={rgb, 255:red, 107; green, 65; blue, 144 }  ,fill opacity=0.55 ] (255,52) -- (265,52) -- (265,62) -- (255,62) -- cycle ;
\draw  [color={rgb, 255:red, 107; green, 65; blue, 144 }  ,draw opacity=1 ][fill={rgb, 255:red, 107; green, 65; blue, 144 }  ,fill opacity=0.55 ] (278,56) -- (288,56) -- (288,66) -- (278,66) -- cycle ;
\draw  [color={rgb, 255:red, 107; green, 65; blue, 144 }  ,draw opacity=1 ][fill={rgb, 255:red, 107; green, 65; blue, 144 }  ,fill opacity=0.55 ] (286,86) -- (296,86) -- (296,96) -- (286,96) -- cycle ;
\draw  [color={rgb, 255:red, 107; green, 65; blue, 144 }  ,draw opacity=1 ][fill={rgb, 255:red, 107; green, 65; blue, 144 }  ,fill opacity=0.55 ] (304,75) -- (314,75) -- (314,85) -- (304,85) -- cycle ;

\draw (49,22) node [anchor=north west][inner sep=0.75pt]   [align=left] {underfitting};
\draw (235,23) node [anchor=north west][inner sep=0.75pt]   [align=left] {optimum};
\draw (410,23) node [anchor=north west][inner sep=0.75pt]   [align=left] {overfitting};

\end{tikzpicture}
}
    \caption{Examples of underfitting, optimum solution, and overfitting in a toy classification problem. The green dots and violet squares represent two classes. The lines represent different models that classify the data. The left plot shows the result of using a model that is too simple or underfitted for the presented dataset, while the right plot shows an overfitted model. }
    \label{fig:classification-problem}
\end{figure*}

During the training procedure, the model can suffer from overfitting and underfitting (see Figure~\ref{fig:classification-problem} for an illustration), which can be described in terms of training and testing errors. 

\textit{Overfitting} is a negative phenomenon that occurs when a learning algorithm generates predictions that fit too closely or exactly to a particular dataset and are therefore not suitable for applying the algorithm to additional data or future observations. In this case, the training error is low but the error computed on a test set is high. The model finds dependencies in the train set which does not hold in the test set. As a result, the model has \textit{high variance}, a problem caused by being highly sensitive to small deviations in the training set.

The opposite of overfitting is \textit{underfitting}, in which the learning algorithm does not provide a sufficiently small average error on the training set. Underfitting occurs when insufficiently complex models are used or the training is stopped too early. In this case, the error is high for both train and test sets. As a result, the model has \textit{high bias}, an error of incorrect assumptions in the learning algorithm. 

The goal is to find the best strategy to reduce overfitting and improve the generalization, or, in other words, reduce the trained model's bias and variance. Ensembles can be used to eliminate high variance and high bias. For example, the \textit{boosting} procedure of several models with high bias can get a model with a reduced bias. In another case, when \textit{bagging}, several low-bias models are connected, and the resulting model can reduce the variance. But in general, reducing one of the adverse effects leads to an increase in the other. This conflict in an attempt to simultaneously minimize bias and variance is called the \textit{bias-variance trade-off}. This trade-off is achieved in the minimum of the test error, see the classical regime region on Figure~\ref{fig:classification-problem}.

\begin{figure*}[ht!]

\begin{center}
\tikzset{every picture/.style={line width=0.75pt}} 

\scalebox{.85}{
\begin{tikzpicture}[x=0.75pt,y=0.75pt,yscale=-1,xscale=1]

\draw [color={rgb, 255:red, 74; green, 74; blue, 74 }  ,draw opacity=1 ] (28.5,245.1) -- (531.5,245.1)(78.8,84) -- (78.8,263) (524.5,240.1) -- (531.5,245.1) -- (524.5,250.1) (73.8,91) -- (78.8,84) -- (83.8,91)  ;
\draw [color={rgb, 255:red, 65; green, 117; blue, 5 }  ,draw opacity=1 ]   (86.89,106) .. controls (154.77,218) and (238.5,226) .. (280.5,161) .. controls (322.5,96) and (326.88,217) .. (458.99,219) ;
\draw [color={rgb, 255:red, 107; green, 65; blue, 144 }  ,draw opacity=1 ]   (87.89,125) .. controls (139,207.46) and (147.54,229.33) .. (278.05,235.05) .. controls (286.82,235.44) and (296.14,235.75) .. (306.06,236) .. controls (426.05,237) and (375.15,237) .. (464.84,238) ;
\draw  [color={rgb, 255:red, 255; green, 255; blue, 255 }  ,draw opacity=1 ][fill={rgb, 255:red, 255; green, 255; blue, 255 }  ,fill opacity=1 ] (24.5,216) -- (77.59,216) -- (77.59,256) -- (24.5,256) -- cycle ;
\draw  [color={rgb, 255:red, 255; green, 255; blue, 255 }  ,draw opacity=1 ][fill={rgb, 255:red, 255; green, 255; blue, 255 }  ,fill opacity=1 ] (35.17,246) -- (120.01,246) -- (120.01,266) -- (35.17,266) -- cycle ;
\draw [color={rgb, 255:red, 74; green, 74; blue, 74 }  ,draw opacity=1 ] [dash pattern={on 0.84pt off 2.51pt}]  (303.18,26) -- (301.97,275.5) ;

\draw (464.43,252) node [anchor=north west][inner sep=0.75pt]  [font=\small,color={rgb, 255:red, 74; green, 74; blue, 74 }  ,opacity=1 ] [align=left] {complexity};
\draw (59.16,64) node [anchor=north west][inner sep=0.75pt]  [font=\small,color={rgb, 255:red, 74; green, 74; blue, 74 }  ,opacity=1 ] [align=left] {error};
\draw (475.9,221) node [anchor=north west][inner sep=0.75pt]  [font=\small,color={rgb, 255:red, 107; green, 65; blue, 144 }  ,opacity=1 ] [align=left] {train };
\draw (457.88,193) node [anchor=north west][inner sep=0.75pt]  [font=\small,color={rgb, 255:red, 65; green, 117; blue, 5 }  ,opacity=1 ] [align=left] {\quad \ test};
\draw (103.43,26) node [anchor=north west][inner sep=0.75pt]  [font=\small,color={rgb, 255:red, 74; green, 74; blue, 74 }  ,opacity=1 ] [align=left] {\begin{minipage}[lt]{123.62pt}\setlength\topsep{0pt}
\begin{center}
classical regime\\(bias-variance trade-off)
\end{center}

\end{minipage}};
\draw (349.13,26) node [anchor=north west][inner sep=0.75pt]  [font=\small,color={rgb, 255:red, 74; green, 74; blue, 74 }  ,opacity=1 ] [align=left] {\begin{minipage}[lt]{84.16pt}\setlength\topsep{0pt}
\begin{center}
modern regime\\(double-descent)
\end{center}

\end{minipage}};

\end{tikzpicture}
}
\end{center}
\caption{Illustration of the double-descent phenomenon.}
\label{fig:double-descent}
\end{figure*}
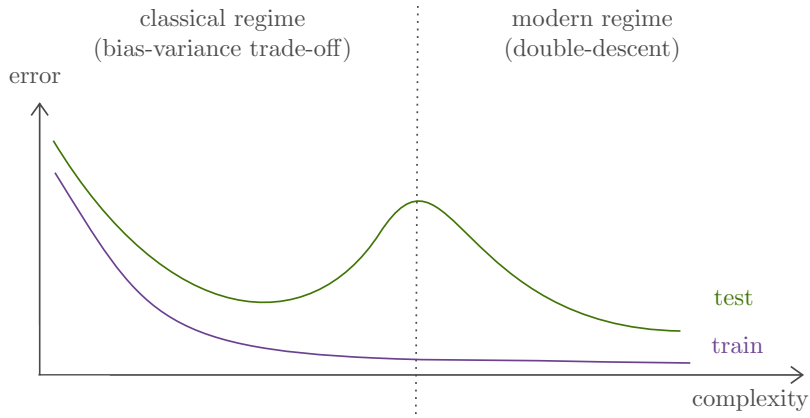

\newparagraph{Modern regime.} 
In the past few years, it was shown that when increasing the model size beyond the number of training examples, the model's test error can start \textit{decreasing again} after reaching the interpolation peak, see Figure~\ref{fig:double-descent}.  
This phenomenon is called \textit{double-descent} by \cite{belkin2019reconciling} who demonstrated it for several machine learning models, including a two-layer neural network.
\cite{nakkiran2021deep} extensively study this double-descent phenomenon for deep neural network models and show the double-descent phenomenon occurs when varying the width of the model or the number of iterations during the optimization. Moreover, the double-descent phenomenon can be observed as a function of dataset size, where more data sometimes lead to worse test performance. It is not fully understood yet why this phenomenon occurs in machine learning models and which inductive biases are responsible for it. However, it is important to take this aspect into account while choosing strategies to improve generalization. 

\newparagraph{Strategies.}
One reason for overfitting is the lack of training data, making the learned distribution not mirror the real underlying distribution. Collecting data arising from all possible parts of the domain to train machine learning models is prohibitively expensive and even impossible. Therefore, enhancing the generalization ability of models is vital in both industry and academic fields. \textit{Data augmentation methods}, which are discussed above in the context of inductive bias, extract more information from the original dataset through augmentations, thus, help to improve the generalization. 

Many strategies for increasing generalization performance focus on the model's architecture itself. Regularization methods are used to encourage a lower complexity of a model. Functional solutions such as dropout regularization~\citep{srivastava2014dropout}, batch normalization~\citep{ioffe2015batch}, transfer learning~\citep{weiss2016survey}, and pretraining~\citep{erhan2010does} have been developed to try to adapt deep learning to applications on smaller datasets. Another approach is to treat the number of training epochs as a hyperparameter and to stop training if the performance of the model on the test dataset starts to degrade, e.g., loss begins to increase or accuracy begins to decrease. This procedure is called \textit{early stopping}.

Though \textit{explicit regularization} techniques are known to improve generalization, their absence does not imply poor generalization performance for deep learning models. Indeed, \cite{zhang2017understanding} argue that neural networks have \textit{implicit regularizations}; for instance, stochastic gradient descent tends to converge to small norm solutions. 
The early stopping procedure can also be viewed as \textit{implicit regularization} method as it implicitly forces to use a smaller network with less \textit{capacity}~\citep{zhang2017understanding,zhang2021understanding}.  

\newparagraph{Generalization bounds.} 
We are often interested in making the discussion on training, validation, and testing sets formal, so as to ensure that our neural network will work well on new data with high probability. We are thus often interested in finding a bound on risk $\mathcal{R}^{\mathcal{L}}_{\mathfrak{D}}(f)=\bE_{(\bx,\by)\sim\mathfrak{D}}\, \mathcal{L}(f(\bx),\by)$ with high probability. 

The most common way of bounding the above in the context of deep neural networks is by use of a test set \citep{langford2005tutorial,kaariainen2005comparison}. One first trains a predictor $f$ using a training set $\mathcal{D}_{\mathrm{train}}$, and then computes a test risk $\mathcal{R}^{\mathcal{L}}_{\mathcal{D}_{\mathrm{test}}}(f)$. For $n_{\mathrm{test}}$ test samples, and in the classification setting, this can be readily turned into a bound on the risk $\mathcal{R}^{\mathcal{L}}_{\mathfrak{D}}(f)$, using a tail bound on the corresponding binomial distribution \citep{langford2005tutorial}. However, this approach has some shortcomings. For one it requires a significant number of samples $n_{\text{test}}$. This can be a problem in that these samples cannot be used for training, possibly hindering the performance of the deep network. At the same time, for a number of fields such as healthcare, the cost of obtaining test samples can be prohibitively high \citep{davenport2019potential}. Finally, even though we can prove that the true risk will be low, we do not get any information about the reason \emph{why} the classifier performs well in the first place.

As such, researchers often use the empirical risk (on the training set) together with the \emph{complexity} \citep{mohri2018foundations} of the classifier to derive bounds roughly of the form 
$$
\mathcal{R}^{\mathcal{L}}_{\mathfrak{D}}(f) \leq \mathcal{R}^{\mathcal{L}}_{\mathcal{D}_{\mathrm{train}}}(f) + \mathrm{complexity}.
$$
Intuitively, the more complex the classifier, the more it is prone to simply memorize the training data, and to learn any discriminative patterns. This leads to high true risk. Traditional data-independent complexity measures such as Rademacher complexity \citep{mohri2018foundations} and VC-dimension \citep{blumer1989learnability} are loose for deep neural networks. This is because they intuitively make a single complexity estimate for the neural network for all possible input datasets. Thus they are pessimistic, as a neural network could memorize one dataset (which is difficult) but learn patterns that generalize on another dataset (which might be easy). 

Based on the above results, researchers focused on complexity measures which are data-dependent \citep{golowich2017size,arora2018stronger,neyshabur2017pac,sokolic2016robust,bartlett2017spectrally,dziugaite2017computing}. This means that they assess the complexity of a deep neural network based on the specific instantiation of the weights that we inferred for a given dataset. The tightest data-dependent generalization bounds are currently PAC-Bayes generalization bounds \citep{mcallester1999some,germain2016pac,dziugaite2017computing,dziugaite2021role}. Contrary to the VC-dimension or the Rademacher complexity, these bounds work for stochastic neural networks (which are also the topic of this review). They can be roughly seen as bounding the mutual information between the training set and the deep neural network weights. The main complexity quantity of interest is typically the \gls{KL} divergence between a prior and a posterior distribution over the deep neural network weights \citep{dziugaite2017computing,mcallester1999some}.
 
 \subsection{Limitations of the frequentist approach to deep learning}\label{sec:frequentist-limitations}
 
Although deep learning models have been largely used in many research areas, such as image analysis ~\citep{Krizhevsky2012}, signal processing~\citep{Graves2013}, or reinforcement learning~\citep{silver2016mastering}, their safety-critical real-world applications remain limited. Here we identify a number of limitations of the frequentist approach to deep learning:
\begin{itemize}
\item miscalibrated and/or overconfident uncertainty estimates~\citep{minderer2021revisiting};
\item non-robustness to \textit{out-of-distribution} samples~\citep{lee2017training,mitros2019validity,hein2019relu,ashukha2020pitfalls}, and sensitivity to \textit{domain shifts}~\citep{ovadia2019can};
\item sensitivity to adversarial attacks by malicious actors~\citep{moosavi2016deepfool,Moosavi-Dezfooli_2017_CVPR,wilson2016deep};
\item poor interpretability of a deep neural networks' inference model~\citep{sundararajan2017axiomatic,selvaraju2017grad,lim2021temporal,koh2017understanding};
\item poor understanding of generalization, over-reliance on validation sets \citep{mcallester1999some,dziugaite2017computing}.
\end{itemize}

\newparagraph{Uncertainty estimates.} We typically distinguish between two types of uncertainty~\citep{der2009aleatory}. \textit{Data (aleatoric) uncertainty} captures noise inherent in the observations. This could be for example sensor noise or motion noise, resulting in uncertainty that cannot be reduced even if more data were to be collected. \textit{Model (epistemic) uncertainty}  derives from the uncertainty on the model parameters, i.e., the weights in case of a neural network~\citep{blundell2015weight}. This uncertainty captures our ignorance about which model generated our collected data. While aleatoric uncertainty remains even for an infinite number of samples, model uncertainty can be explained away given enough data.
For an overview on methods for estimating the uncertainty in deep neural networks see~\cite{gal2016uncertainty,gawlikowski2021survey}.

While \glspl{NN} often achieve high train and test accuracy, the uncertainty of their predictions is miscalibrated \citep{guo2017calibration}.  In particular, in the classification setting, interpreting softmax outputs as per-class probabilities is not well-founded from a statistical perspective. The Bayesian paradigm, by contrast, provides well-founded and well-calibrated uncertainty estimates \citep{kristiadi2020being}, by dealing with stochastic predictors and applying Bayes' rule consistently. 

\newparagraph{Distribution shift.}
Traditional machine learning methods are generally built on the \textit{iid~assumption} that training and testing data are independent and identically distributed. However, the iid~assumption can hardly be satisfied in real scenarios, resulting in uncertainty problems with \textit{in-domain}, \textit{out-of-domain} samples, and \textit{domain shifts}. \textit{In-domain} uncertainty is measured on data taken from the training data distribution, i.e. data from the same domain. \textit{Out-of-domain} uncertainty of the model is measured on data that does not follow the same distribution as the training dataset. Out-of-domain data can include data naturally corrupted with noise or relevant transformations, as well as data corrupted adversarially. Under corruption, the test domain and the training domain differ significantly. However, the model should still not be overconfident in its predictions. 

\cite{hein2019relu} demonstrate that \gls{RELU} networks are always overconfident on out-of-distribution examples: scaling a training point $\bx \in \mathbb{R}^N$ with a scalar $a$ yields predictions of arbitrarily high confidence in the limit $a \to \infty$. \cite{modas2021prime,fawzi2016robustness} discuss that neural networks in the case of classification can suffer from reduced accuracy in the presence of common corruptions. A common remedy is training on  appropriately designed data transformations~\citep{modas2021prime}. However, the Bayesian paradigm should again be beneficial. It is expected that the resulting \emph{Bayesian} neural networks will give more uncertainty in regions far from the training data, thus degrading as images become gradually more corrupted, and diverging from the training data.

\newparagraph{Adversarial robustness.}
As previously mentioned, modern image classifiers achieve high accuracy on iid test sets but are not robust to small, adversarially-chosen perturbations of their inputs. Given an image $\bx$ correctly classified by a neural network, an adversary can usually engineer an adversarial perturbation $\boldsymbol{\delta}$ so small that $\bx + \boldsymbol{\delta}$ looks just like $\bx$ to the human eye, yet the network classifies $\bx + \boldsymbol{\delta}$ as a different, incorrect class. Bayesian neural networks with distributions placed over their weights and biases enable principled quantification of their predictions’ uncertainty. Intuitively, the latter can be used to provide a natural protection against adversarial examples, making BNNs particularly appealing for safety-critical scenarios, in which the safety of the system must be provably guaranteed.

\newparagraph{Interpretability.} Deep neural networks are highly opaque because they cannot produce human-understandable accounts of their reasoning processes or explanations. There is a clear need for deep learning models that offer explanations that users can understand and act upon \citep{lipton2018mythos}. Some models are designed explicitly with interpretability in mind \citep{montavon2018methods,selvaraju2017grad}. At the same time, a number of techniques have been developed to interpret neural network predictions, including among others gradient-based methods \citep{sundararajan2017axiomatic,selvaraju2017grad} which create ``heatmaps'' of the most important features, as well as influence-function-based approaches \citep{koh2017understanding}. The Bayesian paradigm allows for an elegant treatment of interpretability. Defining a prior is central to the Bayesian paradigm, and selecting it helps analyze which tasks are similar to the current task, how to model the task noise, etc.~\citep[see][]{fortuin2021bayesian,fortuin2021priors}. Furthermore, the Bayesian paradigm incorporates a function-space view of predictors \citep{khan2019approximate}. Compared to the weight-space view, this can result in more interpretable architectures.

\newparagraph{Generalization bounds.} It is well known that traditional approaches to proving generalization using generalization bounds fail for deterministic deep neural networks. Such generalization bounds are very useful for cases where we have little training data. In such cases, we might not be able to both train the predictor sufficiently and keep a large enough additional set for validation and testing. Therefore, a generalization bound could ensure that we both train on the full data available while at the same time proving generalization. For example, \cite{zhang2016understanding,golowich2017size} generalization bounds based on Rademacher complexity and the VC dimension provide vacuous bounds on the true error rate (they provide upper bounds larger than 100\%). On the contrary, the Bayesian paradigm currently results in the tightest generalization bounds for deep neural networks, in conjunction with a frequentist approach termed PAC-Bayes \citep{dziugaite2017computing}. Thus following the Bayesian paradigm is a promising direction for tasks with difficult-to-obtain data.\medskip

We introduce the Bayesian paradigm in Section~\ref{section:bml} and then review its application to neural networks in Section~\ref{section:bnn}. 





\newpage 
\section{Bayesian machine learning} 
\label{section:bml}

Achieving a simultaneous design of adaptive and robust systems presents a significant challenge. In their work, \cite{khan2021bayesian} propose that effective algorithms that strike a balance between robustness and adaptivity often exhibit a Bayesian nature, as they can be viewed as approximations of Bayesian inference. The Bayesian approach has long been recognized as a well-established paradigm for working with probabilistic models and addressing uncertainty, particularly in the field of machine learning \citep{ghahramani2015probabilistic}. 
In this section, we will outline the key aspects of the Bayesian paradigm, aiming to provide the necessary technical foundation for the application of Bayesian neural networks.

\subsection{Bayesian paradigm} 
\label{section:bayesian_paradigm}

The fundamental idea behind the Bayesian approach is to quantify the uncertainty in the inference by using probability distributions. Considering parameters as random variables is in contrast to non-Bayesian approaches, also referred to as frequentist or classic, where parameters are assumed to be deterministic quantities. A Bayesian acts by updating their beliefs as data are gathered according to Bayes' rule, an inductive learning process called Bayesian inference. The choice of resorting to Bayes' rule instead of any other has mathematical justifications dating back to works by Cox and by Savage \citep{cox1961algebra,savage1972foundations}.

Recall the following notations: let a dataset $\mathcal{D} = \{(\bx_1, \by_1), \dots, (\bx_n, \by_n)\}$, modeled with a data generating process characterized by a \textit{sampling model} or \textit{likelihood} $p(\mathcal{D} | \bw)$. Let parameters $\bw$ belong to some parameter space denoted by $\boldsymbol{\mathcal{W}}$, usually a subset of the Euclidean space $\mathbb{R}^d$. 
A \textit{prior distribution} $p(\bw)$ represents our prior beliefs about the distribution of the parameters~$\bw$ (more details in Section~\ref{subsec:prior_bayesian}). 
Note that simultaneously specifying a prior $p(\bw)$ and a sampling model $p(\mathcal{D} | \bw)$ amounts to describing the \textit{joint distribution} between parameters $\bw$ and data $\mathcal{D}$, in the form of the product rule of probability $p(\bw, \mathcal{D}) = p(\bw)p(\mathcal{D} | \bw)$. 
The prior and the model are combined with Bayes' rule to yield the \textit{posterior distribution} $p(\bw| \mathcal{D})$ as follows:
\begin{equation}
\label{eq:bayes_theorem}
  p(\bw | \mathcal{D}) = \frac{p(\bw) p(\mathcal{D} | \bw)}{p(\mathcal{D})}.
\end{equation}
The normalizing constant $p(\mathcal{D})$  in Bayes' rule is called the model  \textit{evidence} or \textit{marginal likelihood}. This normalizing constant is irrelevant to the posterior since it does not depend on the parameter $\bw$, which is why Bayes' rule is often written in the form 
\begin{equation*}
    \text{posterior}\propto \text{prior}\times\text{likelihood}.
\end{equation*}
Nevertheless, the model evidence remains critical in \textit{model comparison} and  \textit{model selection}, notably through \textit{Bayes factors}. See for example Chapter 28 in \cite{mackay2003information}, and \cite{lotfi2022bayesian} for a detailed exposition in Bayesian deep learning. It can be computed by integrating over all possible values of~$\bw$:
\begin{equation}
\label{eq:evidence}
  p(\mathcal{D}) = \int p(\mathcal{D} | \bw) p(\bw)  \ddr \bw.
\end{equation}
Using a Bayesian approach, all information conveyed by the data is encoded in the posterior distribution. Often statisticians are asked to communicate scalar summaries in the form of point estimates of the parameters or quantities of interest. A convenient way to proceed for Bayesians is to compute the \textit{posterior mean} of some quantity of interest $f(\bw)$ of the parameters.
The problem therefore comes down to numerical computation of the integral
\begin{equation}
    \label{eq:posterior_mean}
    \mathbb{E}[f(\bw) | \mathcal{D}] = \int f(\bw)p(\bw | \mathcal{D}) \ddr\bw.
\end{equation}
This includes the posterior mean if $f(\bw) = \bw$, as well as \textit{predictive} distributions. More specifically, let $\by^*$ be a new observation associated to some input $\bx^*$ in a regression or classification task; then the prior and posterior predictive distributions are respectively
\begin{align*}
     p(\by^* | \bx^*) &= \mathbb{E}[p(\by^* | \bx^*, \bw)] \\
     &= \int p(\by^* | \bx^*, \bw) p(\bw) \ddr \bw, \\
     \text{and}\quad 
     p(\by^* | \bx^*, \mathcal{D}) &= \mathbb{E}[p(\by^* | \bx^*, \bw) | \mathcal{D}] \\
     &= \int p(\by^* | \bx^*, \bw) p( \bw | \mathcal{D}) \ddr \bw.
\end{align*}
The posterior predictive distribution is typically used in order to assess model fit to the data, by performing posterior predictive checks. More generally, it allows us to account for \textit{model uncertainty}, or \textit{epistemic uncertainty}, in a principled way, by averaging the sampling distribution $p(\by^* | \bx^*, \bw)$ over the posterior distribution $p( \bw | \mathcal{D})$. This model uncertainty is in contrast to the uncertainty associated with data measurement, also called \textit{aleatoric uncertainty} (see Section~\ref{sec:frequentist-limitations}).

\subsection{Priors}
\label{subsec:prior_bayesian}

Bayes' rule \eqref{eq:bayes_theorem} tells us how to update our beliefs, but it does not provide any hint about what those beliefs should be. Often the choice of a prior may be dictated by computational convenience. Let us mention the case of \textit{conjugacy}: a prior is  said to be \textit{conjugate} to a sampling model if the posterior remains in the same parametric family. Classic examples of such conjugate pairs of [prior, model] include the [Gaussian, Gaussian], [beta, binomial], [gamma, Poisson], among others. These three pairs have in common the fact that their model belongs to the exponential family. More generally, any model from the exponential family possesses some conjugate prior. However, the existence of conjugate priors is not a distinguishing feature of the exponential family (for example, the Pareto distribution is a conjugate prior for the uniform model on the interval $[0,\bw]$, for a positive scalar parameter $\bw$). 

Discussing the choice of a prior often comes with the question of \textit{how much information it conveys}? with the distinction of \textit{objective priors} as opposed to \textit{subjective priors}. For example, Jeffreys' prior, defined as being proportional to the square root of the determinant of the Fisher information matrix, is considered an objective prior in the sense that it is invariant to parameterization changes. Uninformative priors often have the troublesome oddity of being \textit{improper}, in the sense of having a density that does not integrate to a finite value (for example, a uniform distribution on an unbounded parameter space). As surprising as it may seem, such priors are commonplace in Bayesian inference and are considered valid ones as soon as they yield a proper posterior, from which one can draw practical conclusions. However, note that an improper prior hinders the use of the prior predictive (which is de facto improper, too), as well as Bayes factors. 
Somehow in the opposite direction to objective priors, subjective priors lie  at the roots of the Bayesian approach, where one's beliefs are encoded through a prior. Eliciting a prior distribution is a delicate issue, see for instance \cite{Mikkola2021} for a recent review.

Critically, encoding prior beliefs becomes more and more difficult with more complex models, where parameters may not have a direct interpretation, and with higher-dimensional parameter spaces, where the design of a prior that adequately covers the space gets intricate. In this case, direct computation of the posterior distribution may become intractable.
If exact Bayesian inference is intractable for a model, its performance hinges critically on the form of approximations made due to computational constraints and the nature of the prior distribution over parameters.

\subsection{Computational methods}
\label{subsec:bayes_comp}

Posterior computation involves three terms: the prior $p(\bw)$, likelihood $p(\mathcal{D} | \bw)$, and  evidence $p(\mathcal{D})$. The evidence integral~\eqref{eq:evidence} is typically not available in closed form and becomes intractable for high-dimensional problems. The impossibility to obtain a precise posterior as a closed-form solution has led to the development of different approximation methods. The inference can be made by considering \textit{sampling strategies} like \gls{MCMC} procedures, or \textit{approximation methods} based on optimization approaches like \textit{variational inference} and the \textit{Laplace method}. 

In recent years, the development of probabilistic programming languages allowed to simplify the implementation of Bayesian models in numerous programming environments: we can mention Stan \citep{carpenter2017stan}, PyMC3 \citep{pymc}, Nimble \citep{de2017programming}, but also some probabilistic extensions of deep learning libraries like TensorFlow Probability \citep{dillon2017tensorflow} and Pyro \citep{bingham2019pyro}, among others.
Nevertheless, there are still many options to be tuned and challenges for each step of a Bayesian model, which we briefly summarize in the following sections. We refer to \cite{gelman2020bayesian} for a detailed overview of the Bayesian workflow.

\subsubsection{Variational inference}\label{sec:VI}

{Variational inference}~\citep{jordan1999introduction,blei2017variational} approximates the true posterior $p(\bw | \mathcal{D})$ with a more tractable distribution $q(\bw)$ called variational posterior distribution. 
More specifically, variational inference hypothesizes an approximation (or variational) family of simple distributions $q$, e.g., isotropic Gaussians, to approximate the posterior: $p(\bw | \mathcal{D}) \approx q(\bw|\theta)$.

Variational inference seeks the distribution parameter $\theta$ in this family by minimizing the \gls{KL} divergence between approximate posteriors and the true posterior.
The \gls{KL} divergence from $q(\cdot | \theta)$ (denoted simply $q$ hereafter) to $p(\cdot | \mathcal{D})$ is defined as
\begin{equation*}
    \mathrm{KL}(q||p(\cdot\vert \mathcal{D})) = \int q(\bw) \log \frac{q(\bw)}{p(\bw | \mathcal{D})} \ddr \bw.
\end{equation*}
Then, Bayesian inference is performed with the intractable posterior  $p(\bw | \mathcal{D})$  replaced by the tractable
variational posterior approximation $q(\bw)$. It is easy to see that
\begin{equation*}
    \mathrm{KL}(q||p(\cdot\vert \mathcal{D})) =  - \int q(\bw) \log \frac{p(\bw) p(\mathcal{D}|\bw) }{q(\bw)} \ddr \bw + \log p(\mathcal{D}).
\end{equation*}
Since the log evidence does not depend on the choice of the approximate posterior $q$, minimizing the \gls{KL} is equivalent to maximizing the so-called \gls{ELBO}:
\begin{align*}
    \mathrm{ELBO}(q) &= \int q(\bw) \log \frac{p(\bw) p(\mathcal{D}|\bw) }{q(\bw)} \ddr \bw \\
    &= 
    -\mathrm{KL}(q||p)    
    +\int q(\bw) \log p(\mathcal{D}|\bw) \ddr \bw.
\end{align*}

To illustrate how to optimize the above objective, let us take the common approach where the prior $p(\bw)$ and posterior $q(\bw)$ are modeled as Gaussians: $p(\bw)=\mathcal{N}(\bw \vert \bw_p , \boldsymbol{\Sigma}_p)$ and  $q(\bw)=\mathcal{N}(\bw \vert \bw_q , \boldsymbol{\Sigma}_q)$, respectively. Then the first term in the \gls{ELBO} can be computed in closed-form by noting that $2\mathrm{KL}(q||p)$ is equal to
%
$$
\mathrm{tr}(\boldsymbol{\Sigma}_p^{-1}\boldsymbol{\Sigma}_p)-d+(\bw_p-\bw_q)^{\top}\boldsymbol{\Sigma}_p^{-1}(\bw_p-\bw_q) + \log \left(\frac{\det \boldsymbol{\Sigma}_p}{\det \boldsymbol{\Sigma}_q} \right),
$$
where $d$ is the dimension of $\bw$. The second term can be approximated through Monte Carlo sampling as 
$$
\int q(\bw) \log p(\mathcal{D}|\bw) \ddr \bw \approx \sum_{i=1}^{S} \log p(\mathcal{D}|\bw_i),
$$
where $\bw_i \sim q(\bw)$, $i=1,\ldots,S$ are Monte Carlo samples. The resulting objective can be typically optimized by gradient descent, by using the reparametrization trick for Gaussians \citep{kingma2015variational}.

\subsubsection{Laplace approximation}\label{sec:BML-laplace}
    
Another popular method is \textit{Laplace approximation} that uses a normal approximation centered at the maximum of the posterior distribution, or maximum a posteriori (MAP). Let us illustrate the Laplace method for approximating a distribution $g$ (typically a posterior distribution) known up to a constant, $g(\bw)=f(\bx)/Z$, defined over a $d$-dimensional space $\boldsymbol{\mathcal{W}}$. At a stationary point $\bw_0$, the gradient $\nabla f(\bw)$  vanishes. Expanding around this stationary point yields
$$
\log f(\bw) \simeq \log f(\bw_0)-\frac{1}{2}(\bw-\bw_0)^{\top}\boldsymbol{\mathrm{A}}(\bw-\bw_0) ,
$$
where the Hessian matrix $\boldsymbol{\mathrm{A}}\in\mathbb{R}^{d\times d}$ is defined by 
$$
\boldsymbol{\mathrm{A}} = -\nabla\nabla\log f(\bw)\vert_{\bw=\bw_0},
$$
and $\nabla$ is the gradient operator. Taking the exponential of both sides we obtain
$$
f(\bw)\simeq f(\bw_0)\exp \left\{-\frac{1}{2}(\bw-\bw_0)^{\top}\boldsymbol{\mathrm{A}}(\bw-\bw_0) \right\}.
$$
The distribution $g(\bw)$ is proportional to $f(\bw)$ and the appropriate normalization coefficient can be found by inspection, giving
\begin{align*}
    g(\bw)&=\frac{\vert \boldsymbol{\mathrm{A}} \vert^{1/2}}{(2\pi)^{d/2}}\exp \left\{-\frac{1}{2}(\bw-\bw_0)^{\top}\boldsymbol{\mathrm{A}}(\bw-\bw_0) \right\}\\
    &=\mathcal{N}(\bw\vert \bw_0,\boldsymbol{\mathrm{A}}^{-1}),
\end{align*}%
where $\vert \boldsymbol{\mathrm{A}} \vert$ denotes the determinant of $\boldsymbol{\mathrm{A}}$. This Gaussian distribution is well-defined provided its precision matrix $\boldsymbol{\mathrm{A}}$ is positive-definite, which implies that the stationary point $\bw_0$ must be a local maximum, not a minimum or a saddle point. 
Identifying $f(\bw)=p(\mathcal{D}\vert\bw)p(\bw)$ and $Z=p(\mathcal{D})$ and applying the above formula results in the typical Laplace approximation to the posterior. To find a maximum $\bw_0$, one can simply run a gradient descent algorithm on $\log f(\bw)=\log p(\mathcal{D}\vert\bw)+ \log p(\bw)$.

\subsubsection{Sampling methods}\label{sec:BML-sampling}

Sampling methods refer to classes of algorithms that use sampling from probability distributions. They are also referred to as \gls{MC} methods when used in order to approximate integrals and have become fundamental in data analysis. In simple cases,  rejection sampling or adaptive rejection sampling can be implemented to return independent samples from a distribution. For more complex distributions, typically multidimensional ones, one can resort to \textit{Markov chain Monte Carlo} (\gls{MCMC}) methods which have become ubiquitous  in Bayesian inference \citep{robert2004monte}. This class of methods consists in devising a Markov chain whose equilibrium distribution is the target posterior distribution. Recording the chain samples, after an exploration phase known as the burn-in period, provides a sample approximately distributed according to the posterior. 

The \gls{MH} method uses some proposal kernel that depends on the previous sample of the chain. \gls{MH} proposes an acceptance/rejection rule for the generated samples. The choice of kernel defines different types of \gls{MH}. For example, random walk \gls{MH} uses a Gaussian kernel with mean at the previous sample and some heuristic variance. In the multidimensional case, Gibbs sampling is a particular case of \gls{MH} when the full-conditional distributions are available. Gibbs sampling is appealing in the sense that samples from the full-conditional distributions are never rejected. However, full-conditional distributions are not always available in closed-form. Another drawback is that the use of full-conditional distributions often results in highly correlated iterations. Many extensions adjust the method to reduce these correlations. 
\gls{MALA} is another special case of \gls{MH} algorithm that proposes new states according to so-called Langevin dynamics. Langevin dynamics evaluate the gradient of the target distribution in such a way that proposed states in \gls{MALA} are more likely to fall in high-probability density regions. 

\gls{HMC} is an improvement over the \gls{MH} algorithm, where the chain's trajectory is based on the Hamiltonian dynamic equations. In Hamilton's equations, there are two parameters that should be computed: a random variable distribution and its moment. Therefore, the exploration space  of a given posterior is expended with its moment. After generating a sample from a given posterior and computing its moment, the stationary principle of Hamilton's equations gives level sets of solutions. \gls{HMC} parameters~--a step size and a number of steps for a numerical integrator~--define how far one should slide the level sets from one space point to the next one in order to generate the next sample.  The \gls{NUTS} is a modification of the original \gls{HMC} which has a criterion to stop the numerical integration. This makes \gls{NUTS} a more automatic algorithm than plain \gls{HMC} because it avoids the need to set the step size and the number of steps.

The main advantage of sampling methods is that they are asymptotically exact: when the number of iterations increases, the Markov chain distribution converges to the (target) posterior distribution. However, constructing efficient sampling procedures with good guarantees of convergence and satisfactory exploration of the sample parameter space can be prohibitively expensive, especially in the case of high dimensions. Note that the initial samples from a chain do not come from the stationary distribution, and should be discarded. The amount of time it takes to reach stationarity is called the mixing time or burn-in time, and reducing it is a key factor for making a sampling algorithm fast. Evaluating convergence of the chain can be done with numerical diagnostics \citep[see for instance][]{gelman1992inference,vehtari2021rank,moins2022local}. 

\subsection{Model selection}
\label{subsec:model_selection}

The Bayesian paradigm provides a principled approach to model selection. 
Let $\{\mathcal{M}_i\}_{i=1}^M$ be a set of $M$ models. We suppose that the data is generated from one of these models but we are uncertain about which one. The uncertainty is expressed through a prior probability distribution $p(\mathcal{M}_i)$ which allows us to express a preference for different models, although a typical assumption is that all models are given equal prior probability~$\nicefrac{1}{M}$. Given a dataset $\mathcal{D}$, we then wish to evaluate the posterior distribution $$p(\mathcal{M}_i|\mathcal{D})\propto p(\mathcal{M}_i)p(\mathcal{D}|\mathcal{M}_i).$$

The \emph{model evidence} $p(\mathcal{D}|\mathcal{M}_i)$ describes the probability that the data were generated from each individual model $\mathcal{M}_i$~\citep{bishop2006pattern}. For a model governed by a set of parameters $\bw$, the model evidence is obtained by integrating out the parameters $\bw$ from the joint distribution $(\mathcal{D},\bw)$, see Equation~\eqref{eq:evidence}:
\begin{align*}
    p(\mathcal{D}|\mathcal{M}_i)&=\int p(\mathcal{D},\bw|\mathcal{M}_i)\mathrm{d}\bw \\
    &=\int p(\mathcal{D}|\bw,\mathcal{M}_i)p(\bw|\mathcal{M}_i)\mathrm{d}\bw.
\end{align*}
The model evidence is also sometimes called the \emph{marginal likelihood} because it can be viewed as a likelihood function over the space of models, in which the parameters have been marginalized out.
From a sampling perspective, the marginal likelihood can be viewed as the probability of generating the dataset $\mathcal{D}$ from a model whose parameters are sampled from the prior. 
If the prior probability over models is uniform, Bayesian \emph{model selection} corresponds to choosing the model with the highest marginal likelihood. The ratio of model evidences $p(\mathcal{D}|\mathcal{M}_i)/p(\mathcal{D}|\mathcal{M}_j)$ for two models is known as a \emph{Bayes factor}~\citep{kass1995bayes}.

The marginal likelihood serves as a criterion for choosing the best model with different hyperparameters.
When derivatives of the marginal likelihood are available (such as for Gaussian process regression), we can learn the optimal hyperparameters for a given marginal likelihood using an optimization procedure.
This procedure, known as \emph{type 2 maximum likelihood}~\citep{bishop2006pattern}, results in the \textit{most likely model} that generated the data. It differs from Bayesian inference which finds the posterior over the parameters for a given model. In the Gaussian process literature, type 2 maximum likelihood optimization often results in better hyperparameters than cross-validation~\citep{lotfi2022bayesian}. For models other than Gaussian processes, one needs to resort to an approximation of the marginal likelihood, typically using the Laplace approximation \citep{bishop2006pattern}.

\newpage 
\section{What are Bayesian neural networks?}
\label{section:bnn}

We have seen now that neural networks are a popular class of models due to their expressivity and generalization abilities, while Bayesian inference is a statistical technique heralded for its adaptivity and robustness.
It is therefore natural to pose the question of whether we can combine these ideas to yield the best of both worlds.
\glspl{BNN} are an attempt at achieving just this.

As outlined in \cref{section:nn}, we aim to infer the parameters of a neural network $\bw \in \boldsymbol{\mathcal{W}}$, which might be the weights and biases of a fully-connected network, the convolutional kernels of a CNN, the recurrent weights of an RNN, etc.
However, in contrast to just using the SGD procedure from \cref{eq:sgd} to get a point estimate for $\bw$, we will try to use the Bayesian strategy from \cref{eq:bayes_theorem} to yield a posterior distribution $p(\bw | \mathcal{D})$ over parameters.
This distribution enables the quantification of uncertainty associated with the model's predictions and can be updated as new data is observed.
While this approach seems straightforward on paper, we will see in the following that it leads to many unique challenges in the context of BNNs, especially when compared to more conventional Bayesian models, such as Gaussian processes \citep{rasmussen2006gaussian}.

Firstly, the weight-space $\boldsymbol{\mathcal{W}}$ of the neural network is often high-dimensional, with modern architectures featuring millions or even billions of parameters.
Moreover, understanding how these weights map to the functions implemented by the network is not trivial.
Both of these properties therefore strongly limit our ability to formulate sensible priors $p(\bw)$, as illustrated in \cref{figure:bnn_visualization_intro}.
We will discuss these challenges as well as strategies to overcome them in more detail in \cref{sec:bnn-priors}, focusing primarily on the theoretical understanding and explanation of empirically observed phenomena, such as the Gaussian process limit in function-space and the relationship between prior selection and implicit and explicit regularization in conventional neural networks.

Secondly, due to the complicated form of the likelihood function (which is parameterized by the neural network itself), neither of the integrals in \cref{eq:evidence} and \cref{eq:posterior_mean} are tractable.
We thus have to resort to approximations, which are again made more cumbersome by the high dimensionality of $\boldsymbol{\mathcal{W}}$.
We will discuss different approximation techniques and their specific implementations in the context of BNNs in \cref{sec:bnn-posteriors}, contrasting their tradeoffs and offering guidance for practitioners.

Whether the aforementioned challenges relating to priors and inference in BNNs are surmountable in practice often depends on the particular learning problem at hand and on the modeling effort and computational resources one is willing to spend.
We will critically reflect on this question in the following and also offer some reconciliation with frequentist approaches later in \cref{section:bayesian_and_non-bayesian}.

\tikzset{%
  input/.style={
      circle,
      draw,
      color={rgb, 255:red, 107; green, 65; blue, 144 },
      draw opacity=1,
      fill={rgb, 255:red, 107; green, 65; blue, 144 },
      fill opacity=0.48,
      minimum size=0.5cm
    },
    every neuron/.style={
      circle,
      draw,
      color={rgb, 255:red, 65; green, 117; blue, 5 },
      draw opacity=1,
      fill={rgb, 255:red, 65; green, 117; blue, 5 },
      fill opacity=0.48,
      minimum size=0.5cm
    },
    neuron missing/.style={
      draw=none, 
      fill=none,
      scale=1,
      text height=0.333cm,
      execute at begin node=\color{black}$\vdots$
    },
    weights/.style={
      draw=none, 
      fill=none,
      scale=1,
      text height=0.333cm,
      execute at begin node=\color{black}$\bw$
    },
    output/.style={
      circle,
      draw,
      color={rgb, 255:red, 107; green, 65; blue, 144 },
      draw opacity=1,
      fill={rgb, 255:red, 107; green, 65; blue, 144 },
      fill opacity=0.48,
      minimum size=0.5cm
    },
}
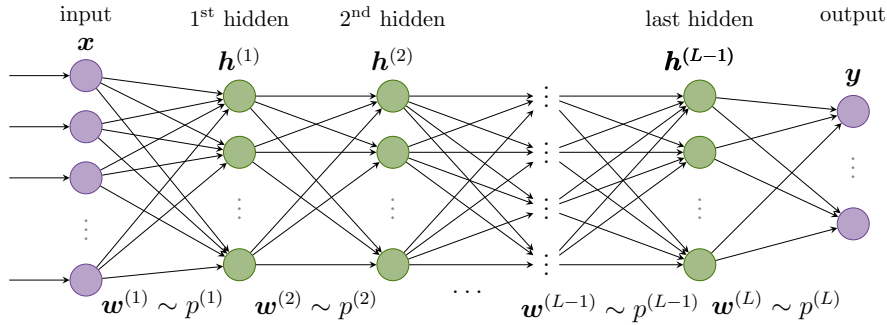
\begin{figure*}[ht!]
\begin{center}
\scalebox{.85}{
\begin{tikzpicture}[x=1.2cm, y=0.8cm, >=stealth]

\foreach \m/\l [count=\y] in {1,2,3,missing,4}
  \node [input/.try, neuron \m/.try] (input-\m) at (0,2.2-\y) {};

\foreach \m [count=\y] in {1,2,missing,3}
  \node [every neuron/.try, neuron \m/.try ] (firsthidden-\m) at (2,1.9-\y*1.1) {};

\foreach \m [count=\y] in {1,2,missing,3}
  \node [every neuron/.try, neuron \m/.try ] (secondhidden-\m) at (4,1.9-\y*1.1) {};

\foreach \m [count=\y] in {1,2,3,4}
  \node [neuron missing/.try, neuron \m/.try ] (thirdhidden-\m) at (6,1.9-\y*1.1) {};

\foreach \m [count=\y] in {1,2,missing,3}
  \node [every neuron/.try, neuron \m/.try ] 
(lasthidden-\m) at (8,1.9-\y*1.1) {};

\foreach \m [count=\y] in {1,missing,2}
  \node [output/.try, neuron \m/.try ] 
(output-\m) at (10,1.6-\y*1.1) {};

\foreach \l [count=\i] in {1,2,3,n}
  \draw [<-] (input-\i) -- ++(-1,0);
  \node [above] at (input-1.north) {$\bx$};
  \node [below] at ([xshift=1.2cm]input-4) {$\boldsymbol{w}^{(1)} \sim p^{(1)}$};

\foreach \l [count=\i] in {1};
\node [above] at (firsthidden-1.north) {$\bh^{(1)}$};
 \node [below] at ([xshift=1.2cm]firsthidden-3.south) {$\boldsymbol{w}^{(2)} \sim p^{(2)}$};

\foreach \l [count=\i] in {1};
  \node [above] at (secondhidden-1.north) {$\bh^{(2)}$};
 \node [below] at ([xshift=1.2cm]secondhidden-3.south) {$\dots$};

\foreach \l [count=\i] in {1}
  \node [above] at (thirdhidden-1.north) {};
 \node [below] at ([xshift=1cm]thirdhidden-4.south) {$\boldsymbol{w}^{(L - 1)} \sim p^{(L - 1)}$};
 
\foreach \l [count=\i] in {1,2,n}
 \node [above] at (lasthidden-1.north) {$\bh^{(L - 1)}$};
 \node [below] at ([xshift=1.2cm]lasthidden-3.south) {$\boldsymbol{w}^{(L)}\sim p^{(L)}$};

\foreach \l [count=\i] in {1}
  \node [above] at (output-1.north) {$\by$};

\foreach \i in {1,...,4}
  \foreach \j in {1,...,3}
    \draw [->] (input-\i) -- (firsthidden-\j);

\foreach \i in {1,...,3}
  \foreach \j in {1,...,3}
    \draw [->] (firsthidden-\i) -- (secondhidden-\j);

\foreach \i in {1,...,3}
  \foreach \j in {1,...,4}
    \draw [->] (secondhidden-\i) -- (thirdhidden-\j);

\foreach \i in {1,...,4}
  \foreach \j in {1,...,3}
    \draw [->] (thirdhidden-\i) -- (lasthidden-\j);

\foreach \i in {1,...,3}
  \foreach \j in {1,...,2}
    \draw [->] (lasthidden-\i) -- (output-\j);

\foreach \l [count=\x from 0] in {\footnotesize{input}, \footnotesize{$1^{\text{st}}$ hidden}, \footnotesize{$2^{\text{nd}}$ hidden}, , \footnotesize{last hidden}, \footnotesize{output}}
  \node [align=center, above] at (\x*2,2) {\l};
\end{tikzpicture}
}
\end{center}
\caption{Bayesian neural network architecture, where weights $\boldsymbol{w}^{(\ell)}$ at layer $\ell$ follow some prior distribution $p^{(\ell)}$.}
\label{figure:bnn_visualization_intro}
\end{figure*}

\subsection{Priors} \label{sec:bnn-priors}

Specifying a prior distribution can be delicate for complex and extremely high-dimensional models such as neural networks. 
Reasoning in terms of parameters is challenging due to their high dimension, limited interpretability, and the over-parameterization of the model. 
Moreover, since the true posterior can rarely be recovered, it is difficult to isolate a prior's influence, even empirically~\citep{wenzel2020good}. This gives rise to the following question: \textit{do the specifics of the prior even matter?} This question is all the more important since inference is usually blunted by posterior approximations and enormous datasets.

The machine learning interpretation of the \textit{no free lunch} theorem states that any supervised learning algorithm includes some \textit{implicit prior} \citep{wolpert1996lack}. From the Bayesian perspective, priors are explicit. Thus, there is an impossibility of the existence of a universal prior valid for any task. This line of reasoning leads to carefully choosing the prior distribution since it can considerably help to improve the performance of the model. 

On the other hand, assigning priors to complex models is often thought of as imposing soft constraints, like regularization, or via data transformations like data augmentation. The idea behind this type of prior is to help and stabilize computation. These priors are sometimes called \textit{weakly informative} or \textit{mildly informative} priors. Moreover, most regularization methods used for point-estimate neural networks can be understood from a Bayesian perspective as setting a prior, see Section~\ref{section:regularization}. 


We review recent works on the influence of the prior in \textit{weight-space}, including how it helps to connect classical and Bayesian approaches applied to deep learning models. 
More discussion on the influence of the prior choice can be found in  \cite{nalisnick2018priors} and \cite{fortuin2021choice}. The choice of the prior and its interaction with the approximate posterior family are studied in \cite{hron2018variational}.

\subsubsection{Weight priors (parameter-space)}

The Gaussian distribution is a common and default choice of prior in Bayesian neural networks. Looking for the \gls{MAP} of such a Bayesian model is equivalent to training a standard neural network under a weighted $\mathscr{L}_2$ regularization (see discussion in Section~\ref{section:regularization}). There is no theoretical evidence that the Gaussian prior is preferable over other prior distribution choices~\citep{murphy2012machine}. Yet, its well-studied mathematical properties lead to having Gaussian distribution as a default prior.   
Further, we review the works that show how different weight priors influence the resulting model. 

\newparagraph{Adversarial robustness and priors.}  In \glspl{BNN}, one can evaluate adversarial robustness with the posterior predictive distribution of the model~\citep{blaas2021effect}. A Lipschitz constant arising from the model can be used in order to quantify this robustness. The posterior predictive depends on the model structure and the weights' prior distribution. In quantifying how the prior distribution influences the Lipschitz constant, \cite{blaas2021effect} establish that for \glspl{BNN} with Gaussian priors, the model's Lipschitz constant is monotonically increasing with respect to the prior variance. It means that lower variance should lead to a lower Lipschitz constant, thus, should lead to higher robustness.

\newparagraph{Gaussian process inducing.} A body of works imposes weight priors so that the induced priors over functions have desired properties, e.g., be close to some \gls{GP}. For instance, \cite{flam2017mapping}, and further extended \cite{flam2018characterizing}, propose to tune priors over weights by minimizing the Kullback--Leibler divergence between \gls{BNN} functional priors and a desired \gls{GP}. However, the Kullback--Leibler divergence is difficult to work with due to the need to estimate an entropy term based on samples. 
To overcome this, \cite{tran2020all} suggest using the Wasserstein distance and provide an extensive study on performance improvements when imposing such priors. Similarly, \cite{matsubara2021ridgelet} use the ridgelet transform~\citep{candes1998ridgelets} to approximate the covariance function of a \gls{GP}.

\newparagraph{Priors based on knowledge about function-space.} Some works suggest how to define priors using information from the function-space since it is easier to reason about than in weight-space. \cite{nalisnick2021predictive} propose \textit{\glsunset{PREDCP}predictive complexity priors} (\glspl{PREDCP}) that constrain the Bayesian prior by comparing the predictions between the model and some less complex reference model. These priors are constructed hierarchically with first-level priors over weights (for example, Gaussian) and second-level hyper-priors over weight priors parameters (for example, over Gaussian variances).
The hyper-priors are defined to encourage functional regularization, e.g., depth selection.

During training, the model sometimes needs to be updated concerning the architecture, training data, or other aspects of the training setup. \cite{khan2021knowledge} propose \textit{knowledge-adaptation priors} (K-priors) to reduce the cost of retraining. The objective function of K-priors combines the weight and function-space divergences to reconstruct past gradients. Such priors can be viewed as a generalization of weight-space priors. More on the function-space priors can be found in the next section.

\subsubsection{Unit priors (function-space)}
\label{section:unit-priors}

Arguably, the prior that matters the most from a practitioner's point of view is the prior induced in function-space, not in parameter space or weight-space~\citep{wilson2020case}. 
The prior seen at the function level can provide insight into what it means in terms of the functions it parametrizes. To some extent, priors on \glspl{BNN}' parameters are often challenging to specify since it is unclear what they actually mean. As a result, researchers typically lack interpretable semantics on what each unit in the network represents. It is also hard to translate some subjective domain knowledge into the neural network parameter priors. Such subjective domain knowledge may include feature sparsity or signal-to-noise ratio \citep[see for instance][]{cui2021informative}. 
A way to address this problem is to study the priors in the function-space, thus raising the natural question: \textit{how to assign a prior on functions of interest for classification or regression settings?}



The priors over parameters can be chosen carefully by reasoning about the functions that these priors induce. Gaussian processes are perfect examples of how this approach works~\citep{rasmussen2006gaussian}. There is a body of work on translating priors on functions given by \glspl{GP} into \gls{BNN} priors~\citep[][]{flam2017mapping,flam2018characterizing,tran2020all,matsubara2021ridgelet}. Recent studies establish a closer connection between infinitely-wide \glspl{BNN} and \glspl{GP} which we review next.


\newparagraph{Infinite-width limit.}
Pioneering work of~\cite{neal1996bayesian} first connected Bayesian neural networks and Gaussian processes. 
Applying the central limit theorem, \citeauthor{neal1996bayesian} showed that the output distribution of a one-hidden-layer neural network converges to a Gaussian process for appropriately scaled weight variances. 
Recently, \cite{matthewsgaussian,lee2018deep} extended Neal's results to deep neural networks showing that their units' distribution converges to a Gaussian process when \textit{the width of all the layers} goes to infinity. 
These observations have recently been significantly generalized to a variety of architectures, including convolutional neural networks~\citep{novak2020bayesian,garriga2019deep}, batch norm and weight-tying in recurrent neural networks~\citep{yang2019tensor}, and ResNets~\citep{hayou2022infinite-depth}. 
There is also a correspondence between \glspl{GP} and models with \textit{attention layers}, i.e., particular layers with an attention mechanism relating different positions of a single sequence to compute a representation of the sequence, see e.g.~\cite{vaswani2017attention}. For multi-head attention architectures, which consist of several attention layers running in parallel, as the number of heads and the number of features tends to infinity, the outputs of an attention model also converge to a \gls{GP}~\citep{hron2020infinite}. Generally, if an architecture can be expressed solely via matrix multiplication and coordinate-wise nonlinearities (i.e., a tensor program), then it has a \gls{GP} limit~\cite{yang2019scaling}.


Further research builds upon the limiting Gaussian process property to devise novel architecture rules for neural networks. Specifically, the \gls{NNGP}~\citep{lee2018deep} describes the prior on function-space that is realized by an iid~prior over the parameters. The function-space prior is a \gls{GP} with a specific kernel defined recursively with respect to the layers. 
For the rectified linear unit (\gls{RELU}) activation function, the Gaussian process covariance function is obtained analytically \citep{cho2009kernel}. Stable distribution priors for weights also lead to stable processes in the infinite-width limit~\citep{favaro2020stable}. 

When the prior over functions behaves like a Gaussian process, the resulting \gls{BNN} posterior in function-space also weakly converges to a Gaussian process, which was firstly empirically shown in~\cite{neal1996bayesian} and \cite{matthewsgaussian} and then theoretically justified by~\cite{hron2020exact}. 
However, given the wide variety of structural assumptions that \gls{GP} kernels can represent~\citep{rasmussen2006gaussian,lloyd2014automatic,sun2018differentiable}, \glspl{BNN} outperform \glspl{GP} by a significant gap in expressive power~\citep{sun2019functional}. 
\cite{adlam2020exploring} show that the resulting \gls{NNGP} is better calibrated than its finite-width analogue.
The downside is its poorer performance in part due to the complexity of training \glspl{GP} with large datasets because of matrix inversions. However, this limiting behavior triggers a new line of research to find better approximation techniques. For example, \cite{yaida2020non} shows that finite-width corrections are beneficial to Bayesian inference. 

Nevertheless, infinite-width neural networks are valuable tools to obtain some theoretical properties on \glspl{BNN} in general and to study the neural networks from a different perspective. It results in learning dynamics via the \textit{neural tangent kernel}~\citep{jacot2018neural}, and an \textit{initialization procedure} via the so-called \textit{Edge of Chaos}~\citep{poole2016exponential,schoenholz2016deep,hayou2019impact}.
We describe below the aforementioned aspects in detail.

\newparagraph{Neural tangent kernel.}
Bayesian inference and the \gls{GP} limit give insights into how well over-parameterized neural networks can generalize. Then, the idea is to apply a similar scheme to neural networks after training and study the dynamics of gradient descent on infinite width. 
For any parameterized function $f(\bx, \bw)$ let:
\begin{equation}
\label{eq:ntk}
K_{\bw}(\bx, \bx') = \langle \nabla_{\bw} f(\bx, \bw),  \nabla_{\bw} f(\bx', \bw) \rangle.
\end{equation}
When $f(\bx, \bw)$ is a feedforward neural network with appropriately scaled parameters, a convergence  $K_{\bw} \to K_\infty$ occurs to some fixed kernel called \gls{NTK} when the network’s widths tend to infinity one by one starting from the first layer~\citep{jacot2018neural}. \cite{yang2019scaling} generalizes the convergence of \gls{NTK} to the case when widths of different layers tend to infinity together. 

If we choose some random weight initialization for a neural network, the initial kernel of this network approaches a deterministic kernel as the width increases. Thus, \gls{NTK} is independent of specific initialization. Moreover, in the infinitely wide regime, \gls{NTK} stays constant over time during optimization. Therefore, this finding enables to study learning dynamics in infinitely wide feed-forward neural networks. For example,  \cite{lee2019wide} show that \glspl{NN} in this regime are simplified to linear models with a fixed kernel. 

While this may seem promising at first, empirical results show that neural networks in this regime perform worse than practical over-parameterized networks~\citep{arora2019exact,lee2020finite}. Nevertheless, this still provides theoretical insight into some aspects of neural network training. 



\newparagraph{Finite width.}
While infinite-width neural networks help derive theoretical insights into deep neural networks, neural networks at finite-width regimes or approximations of infinite-width regimes are the ones that are used in real-world applications. It is still not clear when the  \gls{GP} framework is more amenable to describe the \gls{BNN} behavior. In some cases, finite-width neural networks outperform their infinite-width counterparts~\citep{lee2018deep, garriga2019deep,arora2019exact, lee2020finite}. \cite{arora2019exact} show that convolutional neural networks outperform their corresponding limiting \gls{NTK}. This performance gap is likely due to the finite width effect where a fixed kernel cannot fully describe the \gls{CNN} dynamics. The evolution of the \gls{NTK} along with training has its benefits on generalization as shown in further works~\citep{dyer2020asymptotics,huang2020dynamics}. 

Thus, obtaining a unit prior description for finite-width neural networks is essential. One of the principal obstacles in pursuing this goal is that hidden units in \glspl{BNN} at finite-width regime are dependent~\citep{vladimirova2021dependence}. The induced dependence makes it difficult to analytically obtain distribution expressions for priors in function-space of neural networks. Here, we review works on possible solutions such as the introduction of finite-width corrections to infinite-width models and the derivation of distributional characterizations amenable for neural networks. 

\newparagraph{Corrections.} One of the ways to describe priors in the function-space is to impose corrections to \glspl{BNN} at infinite width. In particular, \cite{antognini2019finite} shows that ensembles of finite one-hidden-layer \glspl{NN} with large width can be described by Gaussian distributions perturbed by a fourth Hermite polynomial. The scale of the perturbations is inversely proportional to the neural network's width. Similar corrections are also proposed in~\cite{naveh2020predicting}. 
Additionally, \cite{dyer2020asymptotics} propose a method using Feynman diagrams to bound the asymptotic behavior of correlation functions in \glspl{NN}. 
The authors present the method as a conjecture and provide empirical evidence on feed-forward and convolutional \glspl{NN} to support their claims.
Further, \cite{yaida2020non} develops the perturbative formalism that captures the flow of pre-activation distributions to deeper layers and studies the finite-width effect on Bayesian inference. 

\newparagraph{Full description.} \cite{springer1970distribution} show that the probability density function of the product of independent normal variables can be expressed through a Meijer G-function. It results in an accurate description of  unit priors induced by Gaussian priors on weights and linear or \gls{RELU} activation functions~\citep{zavatone2021exact,noci2021precise}.  It is the first full description of function-space priors but under strong assumptions, requiring Gaussian priors on weights and linear or \gls{RELU} activation functions, and with fairly convoluted expressions. Though this is an accurate description, it is hard to work with due to its complex structure. 
However, this accurate characterization is in line with works on heavy-tailed properties for hidden units which we discuss further.

\newparagraph{Distributional characteristics.} Concerning the distributional characteristics of neural networks units, a number of alternative analyses to the Gaussian Process limit have been developed in the literature. \cite{bibi2018analytic} provides the expression of the first two moments of the output units of a one-hidden-layer neural network. Obtaining moments is a preliminary step to characterizing a whole distribution. However, the methodology of \cite{bibi2018analytic} is also limited to one-hidden-layer neural networks. Later, \cite{vladimirova2019bayesian,vladimirova2020sub} focuses on the moments of hidden units and shows that moments of any order are finite under mild assumptions on the activation function. More specifically, the \textit{sub-Weibull} property of the unit distributions is shown, indicating that hidden units become heavier-tailed when going \textit{deeper} in the network. 
This result is refined by \cite{vladimirova2021bayesian} who show that hidden units are \textit{Weibull-tail} distributed. Weibull-tail distributions are characterized in a different manner than sub-Weibull distributions, not based on moments but on a precise description of their tails. These tail descriptions reveal differences between hidden units' distributional properties in finite and infinite-width \glspl{BNN}, since they are in contrast with the \gls{GP} limit obtained when going \textit{wider}.



\newparagraph{Representation learning.} The \textit{representation learning} (when the model is provided with data and learned how to represent the features) in finite-width neural networks is not yet well-understood.  However, the infinitely wide case gives rise to studying representation learning from a different perspective. For instance, \cite{zavatone2021asymptotics} compute the leading perturbative finite-width corrections. \cite{aitchison2020bigger} studies the prior over representations in finite and infinite Bayesian neural networks. The narrower, deeper networks, the more flexibility they offer because the covariance of the outputs gradually vanishes as the network size increases. The results are obtained by considering the variability in the top-layer kernel induced by the prior over a finite neural network. 


\subsubsection{Regularization}
\label{section:regularization}

Since deep learning models are over-parametrized, it is essential to avoid overfitting to help these systems generalize well. 
Several explicit regularization strategies are used, including Lasso $\mathscr{L}_1$ and weight-decay $\mathscr{L}_2$ regularization of the parameters. Another way is to inject some stochasticity into the computations that implicitly prevents certain pathological behaviors and thus helps the network to prevent overfitting. The most popular methods in this line of research are dropout~\citep{srivastava2014dropout} and batch normalization~\citep{ioffe2015batch}. It has also been observed that the stochasticity in stochastic gradient descent (which is normally considered as a drawback) can itself serve as an implicit regularizer~\citep{zhang2017understanding}.  

Here we draw connections between popular regularization techniques in neural networks and weight priors in their Bayesian counterparts. \cite{khan2021bayesian,wolinski2020interpreting} have discussed how different regularization methods implicitly correspond to enforcing different priors. 

\newparagraph{Priors as regularization.}
Given a dataset $\mathcal{D} = \{\bx_i, \by_i\}_i$, where $(\bx_i, \by_i)$ are pairs of inputs and outputs,  the \textit{\glsunset{MAP}maximum-a-posteriori} (\gls{MAP}) can be used to obtain point estimation of the parameters:
\begin{align}
    \label{eq:MAP_equation}
 \hat \bw_{\text{MAP}} &= \argmax_{\bw} \log p(\bw | \mathcal{D}) \\
 &= \argmax_{\bw} \left[ \log p(\mathcal{D} | \bw) + \log p(\bw) \right].\nonumber
\end{align}
Performing classification with a softmax link function, $- \log p(\mathcal{D} | \bw)$, corresponds to the cross-entropy loss.
Performing regression with Gaussian noise such that $p(\mathcal{D} \vert \bw) = \prod_{i} p(\by_i | \bw, \bx_i) = \prod_{i} \mathcal{N} \left(\by_i | f(\bx_i, \bw), \sigma^2 \right)$, then $- \log p(\mathcal{D} | \bw)$ is a mean-squared error loss. In this context, the \gls{MAP} estimation with a Gaussian prior $p(\bw)$ is equivalent to optimization of the mean-squared error loss with $\mathscr{L}_2$ regularization, or weight-decay for \glspl{NN}.
Similarly, assigning a Laplace prior to the weights $\bw$ leads to~$\mathscr{L}_1$~regularization. 

In case of a flat prior (uniform and improper) distribution $p(\bw) \propto 1$, the optimization~\eqref{eq:MAP_equation}  boils down to the \textit{\glsunset{MLE}maximum likelihood estimator} (\gls{MLE}):
\begin{equation*}
  \hat \bw_{\text{MLE}} = \argmax_{\bw} \log p(\mathcal{D} | \bw).
\end{equation*}
However, it is important to note that point solutions like $\hat \bw_{\text{MAP}}$ or $\hat \bw_{\text{MLE}}$ are not Bayesian per se, since they do not use \textit{marginalization} with respect to the posterior, a distinguishing property of the Bayesian approach~\citep{wilson2020case}.

\newparagraph{Dropout.} In this regularization technique due to \cite{srivastava2014dropout}, each individual unit is removed with some probability $\rho$ by setting its activation to zero. This can be recast as multiplying the activations $h_{ij}^{(\ell)}$ by a mask variable $m_{ij}^{(\ell)}$, which randomly takes the values 0 or 1: $h_{ij}^{(\ell)} = m_{ij}^{(\ell)} \phi (g_{ij}^{(\ell)})$.
Significant work has focused on the effect of \textit{dropout} as a weight regularizer~\citep{wager2013dropout}. Inductive bias (see Section~\ref{section:inductive_bias}) of dropout was studied in~\cite{mianjy2018implicit}: for single hidden-layer linear neural networks, they show that dropout tends to make the norm of incoming/outgoing weight vectors of all hidden nodes equal.

The dropout technique can be reinterpreted as a form of approximate Bayesian variational inference~\citep{kingma2015variational,gal2016dropout}.  \cite{gal2016dropout} build a connection between dropout and the Gaussian process representation, while \cite{kingma2015variational} propose a way to interpret Gaussian dropout. They develop a \textit{variational dropout} where each weight of a model has its individual dropout rate. \textit{Sparse variational dropout}, proposed by \cite{molchanov2017variational}, extends \textit{variational dropout} to all possible values of dropout rates and leads to a sparse solution. The approximate posterior is chosen to factorize either over rows or over individual entries of the weight matrices. The prior usually factorizes in the same way.  Therefore, performing dropout can be used as a Bayesian approximation. However, as noted by \cite{duvenaud2014avoiding}, dropout has no regularization effect on infinitely-wide hidden layers. 

\cite{nalisnick2019dropout} propose a Bayesian interpretation of regularization via multiplicative noise, with dropout being the particular case of Bernoulli noise. They find that noise applied to hidden units ties the scale parameters in the same way as the \gls{ARD} algorithm~\citep{neal1996bayesian}, a well-studied shrinkage prior. See Section~\ref{sec:BNN-sampling} for more details.

\subsection{Approximate inference for Bayesian neural networks} \label{sec:bnn-posteriors}
\label{section:bnn_posterior}


Exact inference is intractable for Bayesian \glspl{DNN} due to them being highly non-linear functions. Therefore, practitioners resort to approximate inference techniques. Typically, Bayesian approximate inference techniques fall into the following groups: 1) \emph{variational inference},  2) \emph{Laplace approximation}, and 3) \emph{Monte Carlo sampling}. These approaches for \glspl{DNN} have strong similarities to the general approaches described in Section~\ref{subsec:bayes_comp}. However, the following problems arise in the deep learning setting:
\begin{itemize}
    \item Inference is difficult or {intractable}: deep learning models have a very large number of parameters and the training datasets have many samples;  
    \item The \glspl{DNN}' loss landscape is {multimodal}: deep learning models have many local minima with near equivalent training loss.
\end{itemize}
To address these issues, researchers propose more efficient approaches to performing inferences in \glspl{DNN} than those that usually strictly follow the Bayesian paradigm. Depending on one's point of view, these approaches can be seen as either very rough approximations to the true posterior distribution, or as non-Bayesian approaches that still provide useful uncertainty estimates (see more discussion on this in Section~\ref{section:bayesian_and_non-bayesian}). In this section, we give an overview of inference methods in \glspl{DNN} and describe the tractability and multimodality problems in more detail.

\subsubsection{Variational inference}
The first \textit{variational approach} applied to simple neural networks is proposed by \cite{hinton1993keeping}. They use an analytically tractable Gaussian approximation with a diagonal covariance matrix to the true posterior distribution. Further, \cite{barber1998ensemble} show that this approximation can be extended to a general covariance matrix remaining tractable. However, these methods were not deemed fully satisfactory due to their limited practicality. It took eighteen years after the pioneering work of~\cite{hinton1993keeping} to design more practical variational techniques with the work of \cite{graves2011practical} who suggests searching for variational distributions with efficient numerical integration. It allows variational inference for very complex neural networks but remains computationally extremely heavy.
Later, \cite{kingma2014auto} introduce a \textit{reparameterization trick} for the variational evidence lower bound (\gls{ELBO}), yielding a lower bound estimator (see Section~\ref{sec:VI} for a definition of the \gls{ELBO}). This estimator can be straightforwardly optimized using standard stochastic gradient methods.

Along with the advances in variational methods and scalable inference, \cite{blundell2015weight} propose a novel yet efficient algorithm named \textit{\glsunset{BBB}Bayes by Backprop} (\gls{BBB}) to quantify the uncertainty of the neural network weights. It is amenable to backpropagation and returns an approximate posterior distribution, still allowing for complex prior distributions. This method achieves performance on par with neural networks combined with dropout. However, it requires twice more training parameters than the original non-Bayesian neural network due to the need for Gaussian variance parameters. 
At the same time, \cite{hernandez2015probabilistic} suggest the \textit{\glsunset{PBP}probabilistic backpropagation procedure} (\gls{PBP}), which propagates expectations and performs backpropagation in a standard way. In addition, both \gls{BBB} and \gls{PBP} assume independence between weights when optimizing the variational evidence lower bound. While they achieve good results on small datasets, this substantial restrictive assumption on the posterior distribution is likely to result in underestimating the overall posterior uncertainty. 

Variational inference with the \emph{mean-field} assumption~\citep{blundell2015weight,khan2018fast,kingma2015variational,khan2017vprop} achieved early success for \glspl{BNN} due to being computationally cheap and easy to adapt to modern automatic differentiation libraries. However, the mean-field assumption is too restrictive to achieve a reliable posterior approximation.


A whole body of research focuses on adapting variational inference to deep learning models under different optimization methods to find flexible solutions~\citep{louizos2016structured,sun2017learning,osawa2019practical,zhang2018noisy,dusenberry2020efficient,mishkin2018slang}. Typically, more expressive variational posteriors achieve lower test negative log-likelihood and misclassification error, as well as better uncertainty calibration. But variational inference methods are known to suffer from \textit{mode collapse}~\citep{lakshminarayanan2017simple}, i.e., tend to focus on a single mode of the posterior distribution. Thus, the resulting variational posterior distributions still lack expressiveness. Moreover, accurate variational inference for \glspl{DNN} is difficult for practitioners as it often requires tedious optimization of hyperparameters~\citep{wen2018flipout}.

\subsubsection{Laplace approximation}\label{sec:BNN-laplace}

The Laplace approximation can be seen as an intermediate step between variational inference and sampling approaches (see Section~\ref{sec:BML-laplace} for details). It is computationally relatively cheap and useful for theoretical analyses, resulting in an expressive posterior. The main advantage is bypassing the need to optimize the data likelihood of the stochastic predictor. Furthermore, once at a minimum of the loss landscape, Gaussian posteriors can be calculated using simple vector products. It brings significant benefits for \glspl{DNN}, as optimization of the data likelihood for a stochastic neural network is challenging in practice, as we mentioned in the previous section.
 
Works that conventionally popularized \glspl{BNN} are \cite{mackay1992practical} and \cite{neal1992bayesian,neal1996bayesian}. \cite{mackay1992practical} is the first to perform an extensive study using the Laplace method. He experimentally shows that \glspl{BNN} have high predictive uncertainty in the regions outside of the training data. The approach has recently seen a resurgence in interest due to these appealing properties. For a Gaussian posterior, the primary problem is choosing an appropriate approximation to the Hessian (and, therefore, the Gaussian covariance) that is computationally tractable for modern deep networks. \cite{ritter2018scalable} propose the \gls{kfac}  approximation for the Hessian~\citep{martens2015optimizing}. This results in a block diagonal covariance that can be efficiently estimated using the outer products of the gradients. 

\cite{daxberger2021laplace} introduced Laplace Redux, a Python package that automatically computes the Laplace approximation of a given network, for various approximations to the covariance. It has led to a flurry of research on the Laplace approximation that includes works on improving predictions \citep{immer2021improving,antoran2022adapting}, the use the marginal likelihood for model selection \citep{immer2021scalable,lotfi2022bayesian}, as well as learning architectures that are invariant to transformations of the dataset \citep{immer2022invariance}. The Laplace method can also be used to efficiently compute a posterior on a subnetwork, resulting in a more expressive posterior of the whole network~\citep{daxberger2021bayesian}.

\subsubsection{Sampling methods}\label{sec:BNN-sampling}

While the Laplace approximation offers comparable or even better posterior expressiveness and is more stable to optimize than variational inference methods, it still suffers from exploring only a single mode of the loss landscape. Sampling-based approaches offer a potential solution to this problem (see Section~\ref{sec:BML-sampling}). While having a heavy computational burden, they provide (asymptotically) samples from the true posterior and should be able to explore all modes. 

\newparagraph{MCMC/HMC.}
\cite{neal1993bayesian} proposes the first Markov chain Monte Carlo (\gls{MCMC}) sampling algorithm for Bayesian neural networks. He presents \textit{\glsunset{HMC}Hamiltonian Monte Carlo} (\gls{HMC}), a sophisticated gradient-based \gls{MCMC} algorithm. However, \gls{HMC} is prohibitively expensive, requiring full gradient estimates as well as long burn-in periods before providing a single sample from the posterior. Only recently, \cite{izmailov2021bayesian} revisit this approach and apply it to modern deep learning architectures. They use a large number of \glspl{tpu} to perform inference, which is not typically practical. \cite{huang2023efficient} propose a sampling approach based on adaptive importance sampling which exploits some geometric information on the complex (often multimodal) posterior distribution.

\newparagraph{Monte Carlo dropout.}
\cite{gal2016dropout} establish that neural networks with dropout applied before every weight layer are mathematically equivalent to an approximation to the probabilistic deep Gaussian process~\citep{damianou2013deep}. This gives rise to the \gls{MC} dropout method, a prevalent approach to obtaining uncertainty estimates using dropout without additional cost. 
More specifically, the idea of  Monte Carlo dropout is simple and consists of performing random sampling at test time. Instead of turning off the dropout layers at test time (as is usually done), hidden units are randomly dropped out according to a Bernoulli$(p)$ distribution. Repeating this operation $M$ times provides $M$ versions of the MAP estimate of the network parameters $\bw^m$, $m=1,\ldots,M$ (where some units of the MAP are dropped), yielding an approximate posterior predictive in the form of the equal-weight average:
\begin{equation}\label{eq:MCdropout_post_pred}
	p(y\vert x, \mathcal{D})\approx \frac{1}{M}\sum_{m=1}^M p(y\vert x, \bw^m).
\end{equation}
However, the obtained approximate posterior exhibits some pathologies which can result in overconfidence~\citep{foong2019pathologies}.  
Also, Monte Carlo dropout captures some uncertainty from out-of-distribution (OOD) inputs but is nonetheless incapable of providing valid posterior uncertainty. Indeed, Monte Carlo dropout changes the Bayesian model under study, which modifies also the properties of the approximate Bayesian inference performed. Specifically, \cite{folgoc2021mc} show that the Monte Carlo dropout posterior predictive~\eqref{eq:MCdropout_post_pred} assigns zero probability to the true model posterior predictive distribution.


\newparagraph{\gls{sgmcmc}.} 
The seminal work of \cite{welling2011bayesian} combines SGD and Langevin dynamics providing a highly scalable sampling scheme as an efficient alternative to a full evaluation of the gradient. The tractability of gradient mini-batches evaluations in SGD is a common feature behind many subsequent proposals  \citep{ahn2012bayesian,chen2014stochastic,neiswanger2014asymptotically,korattikara2015bayesian,wang2015privacy}.

However, posterior distributions in deep learning often have complex geometries including multimodality, high curvatures, and saddle points. 
The presence of these features heavily impacts the efficacy of SG-MCMC in properly exploring the posterior. 
In order to partially alleviate this problem, \cite{ma2015complete,li2016preconditioned} use adaptive preconditioners to mitigate the rapidly changing curvature. Borrowing ideas from the optimization literature, preconditioners use local information of the posterior geometry at each step to provide more efficient proposals.
To address the multimodality problem, \cite{zhang2019cyclical} propose an SG-MCMC with a cyclical step-size schedule. Alternating large and small step-size proposals, the sampler explores a large portion of the posterior, moving from one mode to another along with a local exploration of each mode. 
Combining these two approaches of adaptive preconditioning and cyclical step-size scheduling yields a state-of-the-art sampling algorithm in Bayesian deep learning~\citep{wenzel2020good}.


Both \gls{MCMC} and stochastic gradient-\gls{MCMC} based methods often result in state-of-the-art results with respect to the test negative log-likelihood error and accuracy \citep{izmailov2021bayesian}, albeit with significant additional computation and storage costs compared to variational inference and the Laplace approximation.





\newpage 
\section{To be Bayesian or not to be?}
\label{section:bayesian_and_non-bayesian}

This section highlights several areas where Bayesian and frequentist approaches overlap, sometimes in a controversial way. In some cases, this overlap brings mutual benefits to both perspectives, resulting in theoretical and empirical advances. However, some topics do not appear to be resolved and remain open for discussion.


In Section~\ref{section:frequentist_bayesian_connections}, we first discuss how the Bayesian framework can lead to insights and improvements for standard \glspl{NN} and vice versa. In Section~\ref{section:initialization}, we describe the connections between randomized initialization schemes for deterministic neural networks and priors in the Bayesian framework. Section~\ref{section:implicit_regularization_sgd} discusses connections between the optimization methods used for deterministic neural networks (such as SGD and ADAM) and posterior distributions in the Bayesian framework. To make \glspl{BNN} competitive with their deterministic counterparts, down-weighting the effect of the prior in approximate inference is often necessary for what is known as \textit{cold} or \textit{tempered} posteriors~\citep{wilson2020case,wenzel2020good}. We discuss this effect and its possible interpretations given in the literature in Section~\ref{section:cold_posterior_effect}. In Section~\ref{section:deep_ensembles}, we discuss the connection between deep ensembles and approximate inference methods.

In Section~\ref{section:performance_certificates}, we discuss certificates that can be obtained for the performance on out-of-sample data for Bayesian neural networks and relate these to the frequentist setting. In Section~\ref{section:frequentist_posterior_validation}, we detail how frequentist guarantees are often used in posterior contraction, showing that the posterior converges to the true posterior when the sample size grows to infinity. In Section~\ref{section:posterior_concentration_generalization}, we describe how PAC-Bayes theorems can be used to certify the performance of  Bayesian neural networks on out-of-sample data with high probability. In Section~\ref{section:marginal_likelihood_generalization}, we discuss the use of the marginal likelihood for model selection. The marginal likelihood has been a subject of debate and various interpretations in recent years, and we detail its connections to frequentist guarantees on out-of-sample performance.   

Finally in Section~\ref{section:benchmarking}, we describe the difficulties encountered when benchmarking Bayesian neural networks. In Section~\ref{section:evaluation_datasets}, we discuss various popular datasets used to evaluate uncertainty in Bayesian deep learning. In Section~\ref{section:evaluation_metrics}, we discuss the different evaluation metrics that are being used for evaluation. Finally in Section~\ref{section:outout_interpretation}, we describe subtle differences in how neural network outputs can be interpreted. These differences can result in different conclusions across different researchers.

\subsection{Frequentist and Bayesian connections}
\label{section:frequentist_bayesian_connections}

Deep neural networks have been typically treated as deterministic predictors. This has been mainly due to the significant computational costs of training. Significant research has been conducted in deriving good initialization schemes for deep neural network parameters and good optimizers. In this section, we explore the connections between the design choices in this frequentist setting and the Bayesian setting. Furthermore, we make connections between deep ensembles and Bayesian inference and provide some possible explanations as to why deterministic neural networks often outperform Bayesian ones.

\begin{mybox}{blue}{TL;DR}
Empirical studies have demonstrated that SGD tends to induce heavy-tailed distributions on the weights of neural networks. This deviates from the prevalent assumption of Gaussian distributions in variational inference. By adopting Bayesian principles, frequentist optimizers can be reinterpreted, leading to enhanced outcomes in uncertainty estimation. However, to achieve competitive performance, it is often necessary to down-weight the influence of the prior distribution. The underlying reasons for this requirement are currently a subject of active debate within the research community. Despite ongoing efforts, Bayesian approaches often struggle to surpass the performance of deep ensembles in various tasks.
\end{mybox}

\subsubsection{Priors and initialization schemes}
\label{section:initialization}

This section reviews techniques for choosing initialization distributions over weights and biases in neural networks. This is by essence a frequentist procedure, but can be interpreted as well as prior elicitation from a Bayesian  standpoint. Initialization schemes often consider Gaussian distributions on the pre-activations. As such they are closely related to the Bayesian wide regime limit when the number of hidden units per layer tends to infinity, because this regime results in a Gaussian process distribution for the weights (Section~\ref{section:unit-priors}). Therefore, approaches to choosing deep neural network initializations should be fruitful in designing better deep neural network priors, and vice versa.

In deep learning, initializing neural networks with appropriate weights is crucial to obtaining convergence. If the weights are too small, then the variance of the input signal is bound to decrease after several layer passes through the network. As a result, the input signal  may drop under some critical minimal value, leading to inefficient learning. On the other hand, if the weights are too large, then the variance of the input signal tends to grow rapidly with each layer. This leads to a saturation of neurons'  activations and to gradients that approach zero. This problem is sometimes referred to as \textit{vanishing gradients}. Opposite to the vanishing problem is accumulating large error gradients during backpropagation. The gradient grows exponentially by repetitively multiplying gradients, leading to \textit{exploding gradients}. So, initialization must help with \textit{vanishing} and \textit{exploding gradients}. 
In addition, the \textit{dying \gls{RELU}} problem is very common when depth increases~\citep{lu2019dying}.

Initialization also must induce \textit{symmetry breaking}, i.e., forcing neurons to learn different functions so that the effectiveness of a neural network is maximized. 
Usually, this issue is solved with the \textit{randomization procedure}. Randomized asymmetric initialization helps to deal with the dying \gls{RELU} problem~\citep{lu2019dying}. 

\cite{frankle2019lottery} proposed an iterative algorithm for parameter pruning in neural networks while saving the original initialization of the weights after pruning, also known as the \textit{winning ticket} of the initialization ``lottery''. Neural networks with such winning tickets could outperform unpruned neural networks; see \cite{malach2020proving} for theoretical investigations. These findings illustrate that neural networks' initialization influences their structure, even without looking like it. 
This also opens a crucial question in deep learning research: \textit{how to best assign network weights before training starts?} 

The standard option for the initialization distribution is independent Gaussian. The Gaussian distribution is easy to specify as it is defined solely in terms of its mean and variance. It is also straightforward to sample from, which is an essential consideration when picking a sampling distribution in practice. In particular, to initialize a neural network, we independently sample each bias~$b_{i}^{(\ell)}$ and each weight~$w_{ij}^{(\ell)}$ from zero-mean Gaussian distributions: 
\begin{equation}
\label{eq:gaussian-init}
    b_{i}^{(\ell)}  \sim \mathcal{N} \left( 0, \sigma^{2}_b \right), \quad w_{ij}^{(\ell)} \sim  \mathcal{N} \left( 0, \frac{\sigma^{2}_w}{H_{\ell - 1}} \right),
\end{equation}
for all $i=1, \dots, H_{\ell}$ and $j=1, \dots, H_{\ell - 1}$. Here, the normalization of weight variances by $1/H_{\ell-1}$ is conventional to avoid the variance explosion in wide neural networks. The bias variance $\sigma^{2}_b$ and weight variance $\sigma_w^{2}$ are called \textit{initialization hyperparameters}. Note that these could depend on the layer index $\ell$.  The next question is \textit{how to set the initialization hyperparameters} so that the output of the neural network is well-behaved.

\newparagraph{Xavier's initialization.} An active line of research studies the propagation of deterministic inputs in neural networks. 
Some heuristics are based on the information obtained before and after backpropagation, such as variance and covariance between the neurons or units corresponding to different inputs. 
\cite{glorot2010understanding} suggest sampling weights from a uniform distribution, saving the variance of activations in the forward and gradients backward passes, which are respectively $1/H_{\ell-1}$ and $1/H_{\ell}$. Since both conditions are incompatible, the initialization variance is a compromise between the two: $2/(H_{\ell-1} + H_{\ell})$. The initialization distribution, called \textit{Xavier's} or \textit{Glorot's}, is the following:
\begin{equation*}
   w_{ij}^{(\ell)} \sim \mathcal{U} \left( -\frac{\sqrt{6}}{\sqrt{H_{\ell-1} + H_{\ell}}}, \frac{\sqrt{6}}{\sqrt{H_{\ell-1} + H_{\ell}}}\right), 
\end{equation*}
with biases $b_i^{(\ell)}$ assigned to zero. The same reasoning can be applied with a zero-mean normal distribution:
\begin{equation*}
   w_{ij}^{(\ell)} \sim \mathcal{N} \left(0, \frac{1}{H_{\ell-1}}\right), \quad \text{or} \quad w_{ij}^{(\ell)} \sim \mathcal{N} \left(0, \frac{2}{H_{\ell-1} + H_{\ell}}\right).
\end{equation*}
This heuristic, based on an analysis of linear neural networks, has been improved by \cite{he2015delving}. First, they show that the variance of the initialization can be indifferently set to $1/H_{\ell-1}$ or $1/H_{\ell}$ (up to a constant factor) without damaging either information propagation or backpropagation, thus making any compromise unnecessary. Second, they show that for the \gls{RELU} activation function, the variance of the Xavier initialization should be multiplied by 2, that is:
\begin{equation*}
   w_{ij}^{(\ell)} \sim \mathcal{N} \left(0, \frac{2}{H_{\ell-1}}\right).
\end{equation*}

\newparagraph{Edge of Chaos.} Other works explore the covariance between pre-activations corresponding to two given different inputs. \cite{poole2016exponential} and  \cite{schoenholz2016deep} obtain recurrence relations by using Gaussian initializations and under the assumption of Gaussian pre-activations. 
They conclude that there is a critical line, so-called \textit{Edge of Chaos}, separating signal propagation into two regions. The first one is an ordered phase in which all inputs end up asymptotically fully correlated, while the second region is a chaotic phase in which all inputs end up asymptotically independent. To propagate the information deeper in a neural network, one should choose initialization hyperparameters $(\sigma^2_b,\sigma_w^2)$ corresponding to the separating Edge of Chaos line, which we describe below in more detail. 

Let $\bx_a$ be a deterministic input vector of a data point $a$, and $g_{i, a}^{(\ell)}$ be the $i$th pre-activation at layer $\ell$ given a data point $a$. Since the weights and biases are randomly initialized according to a centered distribution (some Gaussian), the pre-activations $g_{i, a}^{(\ell)}$ are also random variables, centered and identically distributed.
Let 
\begin{align*}
    q_{aa}^{(\ell)} &= \mathbb{E} \left[ \big(g_{i, a}^{(\ell)} \big)^2\right], \quad 
q_{ab}^{(\ell)} = \mathbb{E} \left[g_{i, a}^{(\ell)} g_{i, b}^{(\ell)}\right], \\
\text{and} \quad
c_{ab}^{(\ell)} &= q_{ab}^{(\ell)}/\sqrt{q_{aa}^{(\ell)} q_{bb}^{(\ell)}},
\end{align*}
be respectively their variance according to input $a$, covariance and correlation according to two inputs $a$ and $b$. 
Assume the Gaussian initialization rules (or priors) of Equation~\eqref{eq:gaussian-init}  for the weights $w_{ij}^{(\ell)}$ and biases $b_i^{(\ell)}$ for all $\ell$, $i$ and $j$, independently. 
Then, under the assumption that pre-activations $g_{i, a}$ and $g_{i, b}$  are Gaussian, the variance and covariance defined above satisfy the following two-way recurrence relations: 
\begin{align*}
    q_{aa}^{(\ell)} &= \sigma_w^2 \int \phi^2\left( u_1^{(\ell-1)} \right) \mathcal{D} g_{i, a} + \sigma_b^2, \\ 
    q_{ab}^{(\ell)} &= \sigma_w^2 \int \phi(u_1^{(\ell-1)}) \phi(u_2^{(\ell-1)}) \mathcal{D} g_{i, a} \mathcal{D} g_{i, b} + \sigma_b^2.
\end{align*}
Here, $\mathcal{D} g_{i, a}$ and  $\mathcal{D} g_{i, b}$ stand for the distributions of standard Gaussian pre-activations $g_{i, a}$ and $g_{i, b}$. Also, $(u_1^{(\ell-1)},u_2^{(\ell-1)})$ correspond to the following change of variables
\begin{align*}
    u_1^{(\ell-1)} &= \sqrt{q_{aa}^{(\ell - 1)}} g_{i, a},\\
    u_2^{(\ell-1)} &= \sqrt{q_{bb}^{(\ell - 1)}} \left(c_{ab}^{(\ell - 1)} g_{i, a} + \sqrt{1 - (c_{ab}^{(\ell - 1)})^2} g_{i, b}\right).
\end{align*}

For any $\sigma_w^2$ and $\sigma_b^2$, there exist limiting points $q^*$ and $c^*$ for the variance, $q^* = \lim_{\ell\to\infty} q_{aa}^{(\ell)}$, and for the correlation, $c^* = \lim_{\ell\to\infty} c_{ab}^{(\ell)}$. Two regions can be defined depending on the value of $c^*$: (i) an \textit{ordered} region if $c^* = 1$, as any two inputs $a$ and $b$, even far from each other, tend to be fully correlated in the deep limit $\ell\to \infty$; (ii) a \textit{chaos} region if $c^* < 1$, as any two inputs $a$ and $b$, even close to each others, tend to decorrelate as $\ell\to \infty$. 

To study whether the point $c^* = 1$ is \textit{stable}, we need to check the values of the derivative: $    \chi_1 = \frac{\partial c_{ab}^{(\ell)}}{\partial c_{ab}^{(\ell - 1)}}\Bigr\vert_{c_{ab}^{(\ell)}=1}$.
There are three cases: (i) \textit{order}, when $\chi_1 < 1$, i.e., the point $c^* = 1$ is stable; (ii)~\textit{transition}, when $\chi_1 = 1$; (iii) \textit{chaos}, when $\chi_1 > 1$, i.e., the point $c^* = 1$ is unstable. Therefore, there exists a separating line in the hyperparameters $(\sigma_w^2,\sigma_b^2)$ space when $c^* = 1$ and $\chi_1 = 1$, that is referred to as \textit{Edge of Chaos}. By assigning the hyperparameters on the Edge of Chaos line, the information propagates as deep as possible from inputs to outputs. 
Note that all of this procedure assumes that pre-activations $g_{i, a}$ and $g_{i, b}$  are Gaussian. 
\cite{wolinski2022imposing} analyze the Edge of Chaos framework without the Gaussian hypothesis.

\subsubsection{Posteriors and optimization methods}
\label{section:implicit_regularization_sgd}

Neural networks without explicit regularization perform well on out-of-sample data~\citep{zhang2017understanding}. This could mean that neural network models, and their architecture or optimization procedure in particular, have an inductive bias which leads to implicit regularization during training. A number of works aim at understanding this topic by analyzing the \gls{SGD} training process.  

One can relate this research direction to the Bayesian perspective. In particular, especially in variational inference, Bayesian practitioners are greatly concerned with the family of posterior distributions they optimize. Insights into the distribution of solutions found by common optimizers could inform the design of better parametric families to optimize. Nevertheless, research on the posterior distributions induced by constant step \gls{SGD} remains in its infancy. Here we review some recent results and argue that it will be fruitful to see their implications for Bayesian inference.

Some works establish that \gls{SGD} induces implicit regularization. For instance, \cite{soudry2018implicit} show that \gls{SGD} leads to $\mathscr{L}_2$ regularization for linear predictors. Further, \gls{SGD} applied to convolutional neural networks of depth $L$ with linear activation function induces $\mathscr{L}_{2/L}$ regularization~\citep{gunasekar2018implicit}.  This type of regularization can be explicitly enforced in the Bayesian setting, for example by the use of an isotropic Gaussian prior. Recent research also proposes that \gls{SGD} induces heavy-tailed distributions in deep neural networks and connects this with compressibility. \cite{mahoney2019traditional} empirically assess the correlation matrix between the weights. Using spectral theory, they show that the correlation matrix converges to a matrix with heavy-tailed entries during training, a phenomenon known as heavy-tailed self-regularization. \cite{gurbuzbalaban2021heavy}  also argue that the gradient noise is heavy-tailed. This has important implications for a Bayesian practitioner. In particular heavy tailedness of the posterior contrasts with the Gaussian distribution assumption typically made in variational inference and the Laplace approximation. Other parametric distributions have been explored in the literature \citep{fortuin2021priors}.

Conversely, different optimizers have been proposed, partly inspired by Bayesian inference \citep{neelakantan2015adding,foret2020sharpness,khan2021bayesian}. \cite{neelakantan2015adding} inject noise into gradient updates, partly inspired by the SGLD algorithm, from Bayesian inference. They show significant improvements in out-of-sample performance. \cite{foret2020sharpness} relax a PAC-Bayesian objective so as to obtain an optimizer called Sharpness Aware Minimizer (SAM). The SAM optimizer makes gradient steps that have been adversarially perturbed so as to improve generalization by converging to flatter minima. SAM significantly improves performance on diverse datasets and architectures. The connections with Bayesian inference are deep; \cite{mollenhoff2022sam} show that SAM is an optimal relaxation of the ELBO objective from variational inference. Finally \cite{mandt2017stochastic} show that \gls{SGD} can be interpreted as performing approximate Bayesian inference.

The line between frequentist and Bayesian approaches is blurred and has been fruitful in both directions. A significant line of works, including \cite{khan2017vprop,khan2018fast,khan2021bayesian,osawa2019practical,mollenhoff2022sam}, explores existing optimizers that work well in the frequentist setting, and reinterprets them as approximate Bayesian algorithms, subsequently proposing novel (Bayesian) optimizers. \cite{khan2018fast} propose a Bayesian reinterpretation of ADAM which has favorable Bayesian inference properties compared to other VI schemes. \cite{mollenhoff2022sam} propose a Bayesian reformulation of SAM which often outperforms the conventional SAM across different metrics. Refer to \cite{khan2021bayesian} for a detailed treatment of this research direction.

\subsubsection{Cold and tempered posteriors}
\label{section:cold_posterior_effect}

A tempered posterior distribution with temperature parameter $T>0$ is defined as $p(\bw|D) \propto \exp(- U (\bw)/T )$, where $U(\bw)$ is the posterior energy function
\begin{equation*}
 U(\bw) \coloneqq - \log p(\mathcal{D} | \bw) - \log p(\bw).
\end{equation*}
Here $p(\bw)$ is a proper prior density function, for example, a Gaussian density. It was recently empirically found that posteriors obtained by exponentiating the posterior to some power greater than one (or, equivalently, dividing the energy function $U(\bw)$ by some temperature $T<1$), performs better than an untempered one, an effect termed the \textit{cold posterior effect} by~\cite{wenzel2020good}.  

The effect is significant for Bayesian inference, as Bayesian inference should in principle result in the most likely parameters given the training data, and thus to optimal predictions. Bayesian inference could be deemed sub-optimal due to the need for cold posteriors, an observation that cannot go unnoticed.  

In order to explain the effect, \cite{wenzel2020good} suggest that Gaussian priors might not be appropriate for Bayesian neural networks, while in other works \cite{adlam2020cold} suggest that misspecification might be the root cause. In some works, data augmentation is argued to be the main reason for this cold posterior effect~\citep{izmailov2021bayesian,nabarro2021data,bachmann2022tempering}: indeed, artificially increasing the number of observed data naturally leads to higher posterior contraction~\citep{izmailov2021bayesian}. 
At the same time, taking into consideration data augmentation does not entirely remove the cold posterior effect for some models. 
In addition, \cite{aitchison2020statistical} demonstrates that the problem might originate in a wrong likelihood specification of the model which does not take into account the fact that common benchmark datasets are highly curated, and thus have low aleatoric uncertainty. 
\cite{nabarro2021data} hypothesize that using an appropriate prior incorporating knowledge of the data augmentation might provide a solution. Finally, heavy-tailed priors such as Laplace and Student-t are shown to mitigate the cold posterior effect~\citep{fortuin2021bayesian}. \cite{kapoor2022uncertainty} argue that for Bayesian classification we typically use a categorical distribution in the likelihood with no mechanism to represent our beliefs about aleatoric uncertainty. This leads to likelihood misspecification. With detailed experiments, \cite{kapoor2022uncertainty} show that correctly modeling aleatoric uncertainty in the likelihood partly (but not completely) alleviates the cold posterior effect. \cite{pitas2022cold} discuss how the commonly used Evidence Lower Bound Objective (a sub-case in the cold posterior effect literature) results in a bound on the KL divergence between the true and the approximate posterior, but not a direct bound on the test misclassification rate. They discuss how some of the tightest PAC-Bayesian generalization bounds (which directly bound the test misclassification rate) naturally incorporate a temperature parameter, that trades off the effect of the prior compared to the training data.

Despite the aforementioned research, the cold and tempered posterior effect has still not been completely explained, posing interesting and fruitful questions for the Bayesian deep learning community.

\subsubsection{Deep ensembles}
\label{section:deep_ensembles}

\cite{lakshminarayanan2017simple} suggest using an \textit{ensemble of networks} for uncertainty estimation,  which does not suffer from mode collapse but is still computationally expensive. Neural network ensembles are multiple MAP estimates of the deep neural network weights. The predictions of these MAP estimates are then averaged to make an ensemble prediction.  Subsequent methods such as \textit{snapshot ensembling}~\citep{huang2017snapshot}, \textit{fast geometric ensembling}~\citep[\acrshort{FGE}:][]{garipov2018loss}, \textit{stochastic weight averaging}~\citep[\acrshort{SWA}:][]{izmailov2019averaging}, \textit{\acrshort{SWA}-Gaussian}~\citep[\acrshort{SWAG}:][]{maddox2019simple}, greatly reduce the computation cost but at the price of a lower predictive performance~\citep{ashukha2020pitfalls}. While \cite{lakshminarayanan2017simple} frame ensemble approaches as an essentially non-Bayesian technique, they can also be cast as a Bayesian model averaging technique~\citep{wilson2020bayesian,pearce2020uncertainty}, and can even asymptotically converge to true posterior samples when adding repulsion~\citep{dangelo2021repulsive}. Specifically they can be seen as performing a very rough Monte Carlo estimate of the posterior distribution over weights. Ensembles are both cheap, but more importantly, typically outperform Bayesian approaches that have been carefully crafted \citep{ashukha2020pitfalls}. This has been empirically explained as resulting from the increased functional diversity of different modes of the loss landscape \citep{fort2019deep}. These are sampled by definition using deep ensembles, and this sampling is hard to beat using Bayesian inference.

\subsection{Performance certificates}    
\label{section:performance_certificates}
\begin{mybox}{blue}{TL;DR}
Bayesian inference is renowned for its ability to provide guarantees on accurate inference of the true posterior distribution given a sufficient amount of data. However, such guarantees pertain to the accurate estimation of the posterior distribution itself, rather than ensuring performance on out-of-sample data. To address the latter, it becomes necessary to rely on generalization bounds, such as the PAC-Bayes framework. Within this framework, model comparison utilizing the marginal likelihood offers guarantees on the performance of the selected model on out-of-sample data, provided that the inference process has been conducted accurately.
\end{mybox}
\subsubsection{Frequentist validation of the posterior}
\label{section:frequentist_posterior_validation}

Recent works address generalization and approximation errors for the estimation of smooth
functions in a nonparametric regression framework using
sparse deep \glspl{NN} and study their posterior mass concentration depending on data sample size.
\cite{schmidt2017nonparametric} shows that sparsely connected deep neural network with \gls{RELU} activation converges at near-minimax rates when estimating H\"older-smooth functions, preventing the curse of dimensionality. Based on this work, \cite{rockova2018posterior} introduce a Spike-and-Slab prior for deep \gls{RELU} networks which induces a specific regularization scheme in the model training. The obtained posterior in such neural networks concentrates around smooth functions with near-minimax rates of convergence. 
Further, \cite{kohler2021rate} extend the consistency guarantees for H\"older-smooth functions of \cite{schmidt2017nonparametric} and \cite{rockova2018posterior} to fully connected neural networks without the sparsity assumption. Alternatively, \cite{suzuki2018adaptivity} provides generalization error bounds for more general functions in Besov spaces and variants with mixed smoothness.

One of the ways to visualize the obtained uncertainty is using credible sets around some parameter estimator, where the credible region contains a large fraction of the posterior mass~\citep{szabo2015frequentist}. \cite{hadji2021can} study the uncertainty resulting from using Gaussian process priors. 
\cite{franssen2022uncertaintyBNN} provide Bayesian credible sets with frequentist coverage guarantees for standard neural networks trained with gradient descent. Only the last layer is assigned a prior distribution on the parameters and the output obtained from the previous layer is used to compute the posterior. 


\subsubsection{Posterior concentration and generalization to out-of-sample data}
\label{section:posterior_concentration_generalization}

It is interesting to take a step back and evaluate the difference in \emph{goals} between the frequentist and Bayesian approaches to machine learning. The Bayesian approach emphasizes that the posterior concentrates around the true parameter as we increase the training set size, see the previous section. The  primary goal of the frequentist approach is the performance on out-of-sample data, i.e., generalization, see Section~\ref{section:generalization_and_overfitting}. This performance is quantified with validation and test sets. These two goals frequently align, although posterior concentration guarantees and performance on out-of-sample data are typically not mathematically equivalent problems. 
    
When the number of parameters is  smaller than the number of samples $n$, typically in parametric models, the posterior  concentrates on the true set of parameters when $n$ approaches to infinity. In such cases, the posterior tends to a Dirac delta mass centered on the true parameters. In this setting, we can then argue that we are making predictions using the true predictive distribution, and frequentist and Bayesian goals align. We have inferred the true predictor (according to Bayesian goals) and can be sure that we cannot improve the predictor loss on new out-of-sample data, such as validation and test sets (according to the frequentist approach priorities).
    
However, neural networks do not operate in this regime. They are heavily overparametrized, so that Bayesian model averaging always occurs empirically. Usually, we are not interested in the proposed model itself but in its predictions based on new data. Also, due to misspecification, we cannot even assume that we are concentrating around the true predictor. At this point, the frequentist and Bayesian goals diverge. But it is clear that in a non-asymptotic setting and where performance on out-of-sample data is crucial, we need a more detailed description of the predictor's loss on new data.
    
One way to approach this problem is through generalization bounds~\citep{vapnik1999overview} which directly link the empirical loss on the training set with the loss on new data. Of particular interest are PAC-Bayes generalization bounds \citep{mcallester1999some,germain2016pac,dziugaite2017computing,dziugaite2021role}, which directly bound the true risk of a stochastic predictor. Minimizing the \gls{ELBO} objective in variational inference corresponds to minimizing a  PAC-Bayes bound \citep{dziugaite2017computing}, and thus a bound on the true risk. If alternatively one samples \emph{exactly} from the Gibbs posterior (for example using \gls{MCMC}), then one is still minimizing a  PAC-Bayes bound on the true risk \citep{germain2016pac}. Furthermore, in this setting, maximizing the \emph{marginal likelihood} of the model is equivalent to minimizing a  PAC-Bayes bound \citep{germain2016pac} and it has been shown that PAC-Bayes bounds can be used to meta-learn better priors for BNNs~\citep{rothfuss2021pacoh, rothfuss2022pac}.
    
Of particular interest in this discussion is that performing Bayesian inference is equivalent to minimizing \emph{some}  PAC-Bayes bound and not necessarily \emph{the tightest} bound.  PAC-Bayes bounds typically include a temperature parameter that trades-off the empirical risk with the KL complexity term, and plays a crucial role in the bound tightness (see Section~\ref{section:cold_posterior_effect}). An interesting open question is whether this temperature parameter provides a justification for the \emph{cold posterior effect}, with a number of works providing evidence to support this view \citep{grunwald2012safe,pitas2022cold}.

\subsubsection{Marginal likelihood and generalization}
\label{section:marginal_likelihood_generalization}

The marginal likelihood~\citep{mackay2003information} has been explored for model selection, architecture search and hyperparameter learning for deep neural networks. While estimating the marginal likelihood and computing its gradients is relatively straightforward for simple models such as Gaussian processes~\citep{bishop2006pattern}, deep neural networks often require to resort to approximations. 

One approach is the Laplace approximation as previously discussed in Section~\ref{sec:BNN-laplace}.  \cite{daxberger2021laplace,immer2021scalable,immer2022invariance} use the Laplace approximation to the marginal likelihood to select the best-performing model on out-of-sample data. They also use the marginal likelihood to learn hyperparameters, in particular the prior variance and the softmax temperature parameter. For the case of the Laplace approximation, the marginal likelihood of training data $\mathcal{D}$ given the deep neural network architecture  $\mathcal{M}$ can be written as 
\begin{align}
    \log p(\mathcal{D}\vert\mathcal{M})&=\log p(\mathcal{D}\vert \hat \bw_{\text{MAP}},\mathcal{M})+\log p(\hat \bw_{\text{MAP}}\vert\mathcal{M})\nonumber\\
    &+\frac{d}{2}\log 2\pi +\frac{1}{2} \log \left\vert \bLambda_{\hat \bw_{\text{MAP}}} \right\vert,
\end{align}
where  $d$ is the number of weights of the neural network, $\hat \bw_{\text{MAP}}$ is a MAP estimate of the network parameters, and $\bLambda_{\hat \bw_{\text{MAP}}}$ is the precision matrix of the Gaussian posterior distribution under the Laplace approximation. Similarly to the discussion in Section~\ref{sec:BNN-laplace}, the primary computational problem is forming the precision matrix and estimating its determinant. Again the generalized Gauss--Newton approximation and the Empirical Fisher approximation to the Hessian (and correspondingly to the precision matrix) are the most common and efficient approximations, and are the ones used in~\cite{daxberger2021laplace,immer2021scalable}. On a conceptual level, a main criticism of the Laplace approximation for the marginal likelihood of deep neural networks is that it is unimodal while the loss landscape of deep neural networks has multiple minima \citep{lotfi2022bayesian}. This might severely underestimate the volume of good solutions with respect to bad solutions given the prior, which is essentially what the marginal likelihood estimates. A further criticism is that this approximation to the marginal likelihood is sensitive to the prior variance. Indeed for a fixed prior variance across different neural network architectures, \cite{lotfi2022bayesian} show that the marginal likelihood performs poorly for model selection. However optimizing a common prior covariance across layers, or optimizing different prior variances for different layers, results in a better empirical correlation of the marginal likelihood with out-of-sample performance. Overall, the marginal likelihood provides reasonable predictive power for out-of-sample performance  for deep neural networks, and as such constitutes a reasonable approach to model selection.

A different approach is to resort to the product decomposition of the marginal likelihood as 
\begin{align} \label{CLML_eq}
    \log p(\mathcal{D}\vert\mathcal{M})&=\log \prod_{i=1}^n p(\mathcal{D}_i\vert\mathcal{D}_{<i},\mathcal{M})\\ 
    &= \sum_{i=1}^n \log [ \bE_{p(\theta|\mathcal{D}_{<i})} p(\mathcal{D}_i\vert\theta,\mathcal{M}) ]\nonumber
\end{align}
which measures how good the model is at predicting each data point $\mathcal{D}_i$ in sequence given every data point before it, $\mathcal{D}_{<i}$. Based on this observation, \cite{lyle2020bayesian,ru2021speedy} propose the sum of losses of the different batches across an epoch as an approximation to the marginal likelihood. Then, they use this as a measure of the ability of a model to generalize to out-of-sample data. They also propose different heuristics, such as taking the average of the sum of the losses over multiple epochs. A further heuristic is keeping only the last epochs of training while rejecting the sum of the losses of the first epochs. Finally the authors propose to train the neural network for a limited number of epochs, for example only half of the number of epochs that would be typically used to train to convergence. As such the approach is computationally efficient, requiring only partial convergence of the deep neural network and a calculation of the training losses over batches, which are efficient to estimate. 

\cite{ru2021speedy} compare their approach to the task of architecture search to other common approaches. These approaches are a mixture of heuristics and frequentist statistics. The first is the sum of validation losses up to a given epoch. The second is the validation accuracy at an early epoch, which corresponds to the early-stopping practice whereby the user estimates the final test performance of a network using its validation accuracy at an early epoch. The third is the learning curve extrapolation method, which was proposed in \cite{baker2017accelerating} and which trains a regression model on previously evaluated architecture data to predict the final test accuracy of new architectures. The inputs for the regression model comprise architecture meta-features and learning curve features up to a given epoch. They also compare to zero-cost baselines: an estimator based on input Jacobian covariance~\citep[JavCov,][]{mellor2021neural} and two adapted from pruning techniques~\citep[SNIP and SynFlow,][]{abdelfattah2021zero}. The authors demonstrate significantly better rank-correlation in neural architecture search (NAS) for the marginal likelihood approach compared to the baselines. These results have been further validated in \cite{lotfi2022bayesian}. 

\cite{ru2021speedy} have however been criticized for using the term ``training speed'' (as in the number of steps needed to reach a certain training error) to describe their approach. In short, they claim that Equation~\eqref{CLML_eq} corresponds to some measure of training speed, and thus they claim that \emph{training faster corresponds to better generalization}. This however is not generally true as pointed out in \cite{lotfi2022bayesian}. The marginal likelihood can be \emph{larger} for a model that converges \emph{in more steps} (than another model) if the marginal likelihood at step $i=1$ in decomposition~\eqref{CLML_eq} is higher.  

There is a debate as to whether the marginal likelihood is appropriate for model selection at all. \cite{lotfi2022bayesian}  make a distinction between the question “what is the probability that a prior model generated the training data?” and the question “how likely is the posterior, conditioned on the training data, to have generated withheld points drawn from the same distribution?”. They claim that the marginal likelihood answers the first question and not the second. However, high marginal likelihood also provides frequentist guarantees on \textit{out-of-sample} performance through PAC-Bayesian theorems~\citep{germain2016pac}. 
If one selects a model based on the marginal likelihood and also performs Bayesian inference correctly, then the resulting model and its posterior over parameters are guaranteed to result in good performance on out-of-sample data. Overall, the debate is far from concluded, and in light of the good empirical performance of the marginal likelihood, more research is warranted in its direction. 


\subsection{Benchmarking}
\label{section:benchmarking}

\glspl{BNN} present unique challenges in terms of their evaluation and benchmarking. Two main challenges are the choice of the evaluation \emph{datasets} and \emph{metrics} that do not have a consensus in the society.
Non-consensus reflects a difficulty with clearly defining the goals of Bayesian deep learning in a field traditionally viewed through a \emph{frequentist} lens, and more specifically through performance on out-of-sample data. 

\begin{mybox}{blue}{TL;DR}
When evaluating Bayesian deep learning approaches, it is customary to employ standard datasets like MNIST, FMNIST, and CIFAR-10 as benchmark datasets. In contrast, the adoption of Imagenet is relatively infrequent due to computational limitations associated with its larger scale. The primary advantage of Bayesian methods lies in their ability to provide reliable calibration metrics, such as the Expected Calibration Error (ECE) and the Thresholded Adaptive Calibration Error (TACE). However, caution must be exercised when comparing models that utilize temperature scaling, as different practitioners may perceive it either as an integral component of the model or as part of the evaluation metric.
\end{mybox}


\subsubsection{Evaluation datasets}
\label{section:evaluation_datasets}

There is no single dataset benchmark specifically tailored to Bayesian deep learning models. The significant computational burden imposed by Bayesian deep learning has limited the dataset choice to smaller datasets comparatively with how deterministic neural networks are evaluated. 
The majority of papers, for example \cite{khan2018fast,khan2021knowledge,maddox2019simple,gal2016dropout,izmailov2021bayesian,wenzel2020good}, use the {MNIST}, {CIFAR-10}, {CIFAR-100}, and  {Imagenet} datasets combined with different metrics. Also popular are the {UCI} regression datasets~\citep{Dua:2019} such as Concrete, Energy, Kin8nm, Naval, Power, Wine, Yacht.
However, these datasets are smaller than what would be a typical benchmark for deep learning models such as modern \glspl{CNN} architectures. 
    
The Imagenet dataset is notably absent from most papers, as benchmarking on Imagenet for Bayesian deep learning is computationally prohibitive for most practitioners. Both the CIFAR-10 dataset and the CIFAR-100 dataset have 50K training samples, while Imagenet has 1.3M training samples. Correspondingly, a \gls{RESNET} for CIFAR-10 might typically have 20 layers while 50 or more layers are typically required for Imagenet. Many Bayesian deep learning approaches incur linear increases in computational cost with respect to the training set size when estimating the Hessian, and quadratic increases in storage cost with respect to the number of parameters, potentially resulting in prohibitive costs.
    
There have been some efforts to create a standardized benchmarking task that reflects the complexities and challenges of safety-critical real-world tasks while adequately accounting for the reliability of the models predictive uncertainty estimates. \cite{band2021benchmarking} undertake this issue applying it to the \emph{diabetic retinopathy} dataset. 
The dataset is made of retina images associated with diabetic retinopathy, a medical condition considered as the leading cause of vision impairment and blindness. 
Unlike in other works on diabetic retinopathy detection, the benchmarking tasks presented in \cite{band2021benchmarking}  are specifically designed to assess the reliability of machine learning models and the quality of their predictive uncertainty estimates using both aleatoric and epistemic uncertainty estimates. Examples of tasks are \emph{selective prediction} and \emph{expert referral}. These mirror real-world scenarios of predictive uncertainty estimates to identify data points where the likelihood of an incorrect prediction is particularly high and refer for further review to experts, in this case, doctors. Better uncertainty estimates would lead to lower error rates on this task. However, the work of \cite{band2021benchmarking} has not been widely adopted.

\subsubsection{Evaluation metrics-tasks} 
\label{section:evaluation_metrics}
   
For most popular machine learning tasks, the community has reached a consensus on the appropriate evaluation metric of choice, such as the \gls{MSE} for regression and zero-one loss for classification. In the case of Bayesian deep learning, there is not yet a clear choice. Should the Bayesian approach improve on frequentist metrics such as misclassification rate on held-out data? Should it provide solutions to known issues of traditional approaches, such as improved robustness to adversarial and non-adversarial noise? Or should Bayesian approaches be evaluated on different metrics altogether or on metrics that capture \emph{uncertainty}?

\newparagraph{Standard losses.} Practitioners propose several metrics (and corresponding tasks) for the evaluation of Bayesian deep learning approaches. By far the most popular choice is to evaluate frequentist metrics on held-out data that are the \gls{MSE} for regression and the zero-one loss for classification~\citep{khan2018fast,khan2021knowledge,gal2016dropout,izmailov2021bayesian,wenzel2020good}. The intuition behind this choice is that the posterior predictive distribution should improve upon deterministic predictions as multiple predictions from the posterior are averaged. For example, in the case of classification, the posterior predictive is meant to better approximate the \emph{probability} that a given class is correct.

One problem with this approach is that Bayesian approaches have typically provided inconsistent gains for this task-metric combination. For example, sometimes Bayesian approaches improve upon a deterministic neural network and sometimes provide worse results. See for example Figure 5 in  \cite{izmailov2021bayesian} where the \gls{MSE} is evaluated on UCI regression tasks. Similarly, Figure 4.a. in \cite{daxberger2021laplace} shows that the Laplace approximation to a \gls{DNN} posterior does not improve upon the \gls{MAP} solution.  

\cite{wenzel2020good} point out that one can improve upon deterministic neural networks by using heuristics such as cold posteriors which however deviate from the Bayesian paradigm. One common switch away from \gls{MSE} and zero-one loss consists in evaluating the (negative) log-likelihood of the test data. Here, Bayesian approaches often outperform  frequentist ones, but exceptions remain \citep{wenzel2020good}.

\newparagraph{Calibration.} 
By far, the metric on which Bayesian neural networks consistently outperform is the \emph{calibration} for a classification task, i.e., if a classifier has $x\%$ confidence when classifying samples from a sample set, it should also be correct $x\%$ of the time. 
The two most popular metrics for evaluating calibration are the \emph{expected calibration error} \citep[ECE:][]{degroot1983comparison}, and the \emph{thresholded adaptive calibration error} \citep[TACE:][]{nixon2019measuring}. For this type of task-metric combination Bayesian and Bayesian-like approaches such as ensembles (see Section~\ref{section:deep_ensembles}) consistently outperform deterministic neural networks~\citep{izmailov2021bayesian,daxberger2021laplace,maddox2019simple}. \cite{ashukha2020pitfalls} provide a detailed discussion on evaluation metrics for uncertainty estimation as well as common pitfalls. They argue that for a given metric one should always compare a Bayesian method to an ensemble. Ensembles provide good gains in different uncertainty metrics for each new ensemble member. Bayesian methods, often do not result in the same gains for each new sample from the posterior.
    
Other methods for evaluating calibration include reliability diagrams~\citep{vaicenavicius2019evaluating} and calibration curves~\citep{maddox2019simple}. A strength of these metrics is that they are generally clear, direct and intuitive. One weakness of them is that like other visual methods, they are subject to misinterpretation. For example, calibration curves provide a simple and intuitive way to determine which classifier is better calibrated than others when the difference between the classifiers is large. However, when the difference is small or the classifier is miscalibrated only for certain confidence levels, then deriving reliable conclusions becomes more tedious.
One caveat is that a classifier that is guessing completely at random and assigns the marginal class frequencies as predictive probabilities to each data point would trivially achieve a perfect ECE of 0~\citep{gruber2022better}.
Moreover, it has been argued that while many of these metrics measure marginal uncertainties on single data points, joint uncertainties across many points might be more relevant in practice, e.g., for sequential decision-making~\citep{osband2022neural}.
    
\newparagraph{Robustness.} 
There are many works that explored robustness to adversarial noise~\citep{louizos2017multiplicative,rawat2017adversarial,liu2018adv,grosse2018limitations,bekasov2018bayesian} and to non-adversarial noise~\citep{gal2016dropout,daxberger2021bayesian,dusenberry2020efficient,daxberger2021laplace,izmailov2021dangers}, including Gaussian noise, image rotations, among others. \cite{band2021benchmarking} analyze a form of distribution shift whereby classifiers are trained on a set of images for which diabetic retinopathy exists at moderate levels. Then, the evaluation of the classifiers is assessed on a test set where diabetic retinopathy is more severe. The intuition is that Bayesian approaches should correctly classify these corrupted samples and assign low confidence in their predictions. The results in these tasks-metrics are mixed. In the adversarial setting, \glspl{BNN} are typically far from the state-of-the-art defenses against adversarial attacks. In the non-adversarial setting, some works show \emph{improved} robustness \citep{daxberger2021bayesian}, while others show \emph{reduced} robustness \citep{izmailov2021dangers}.
    

\subsubsection{Output interpretation}
\label{section:outout_interpretation}


We conclude by analyzing the output of \glspl{BNN} with the question of its probabilistic interpretation and its relation to evaluation metrics. We restrict the discussion to classification models, though the discussion for other tasks is similar. 
Both frequentist and Bayesian practitioners recognize that the outputs of a deep neural network classifier often do not accurately reflect the probability of choosing the correct class. That is, the \glspl{NN} are not well calibrated. However, frequentist and Bayesian communities propose different solutions. The frequentist solution is to transform the outputs of the classifier through a post-processing step to obtain well-calibrated outputs. Common approaches include \emph{histogram binning} \citep{zadrozny2001obtaining}, \emph{isotonic regression} \citep{zadrozny2002transforming}, \emph{Bayesian binning into quantiles} \citep{naeini2015obtaining} as well as \emph{Platt scaling} \citep{platt1999probabilistic}. 

In a Bayesian setting, the predictive distribution has a clear interpretation, that is the confidence of the model in each class for a given input signal. Confusion can arise from the fact that scaling is sometimes considered part of an evaluation metric. For example, \cite{guo2017calibration} consider \emph{Platt scaling} as a post-processing step (therefore it defines a new model), while  \cite{ashukha2020pitfalls} propose that it be incorporated into a new evaluation metric. The choice of which of the two is true is important as the impact of recalibration methods can be significant in improving the calibration of a model. Thus, if one considers recalibration as defining a new model as in \cite{ashukha2020pitfalls}, then a \gls{kfac} Laplace \gls{BNN} outperforms its corresponding frequentist one significantly in calibration. If recalibration is part of the evaluation metric, then the gains become marginal.

\newpage 
\section{Conclusion}
\label{section:conlusion}




The present review encompasses various topics, such as the selection of prior (Section~\ref{subsec:prior_bayesian}), computational methods (Section~\ref{subsec:bayes_comp}), and model selection (Section~\ref{subsec:model_selection}), which pertain to Bayesian problems in a general sense as well as Bayesian neural networks specifically. This comprehensive perspective enables the contextualization of the diverse inquiries that emerge within the Bayesian deep learning community.

Despite the growing interest and advancements in inference techniques for Bayesian deep learning models, the considerable computational burden associated with Bayesian deep learning approaches remains a primary hindrance. Consequently, the community dedicated to Bayesian deep learning remains relatively small, and the adoption of these approaches in the industry remains limited.

The establishment of a consensus regarding evaluation metrics and benchmarking datasets for Bayesian deep learning has yet to be attained. The lack of consensus stems from the challenge of precisely defining the objectives of Bayesian deep learning within a domain traditionally perceived through a \emph{frequentist} framework, particularly emphasizing performance on out-of-sample data.


This review provides readers with a thorough exposition of the challenges intrinsic to Bayesian deep learning, while also shedding light on avenues that warrant additional exploration and enhancement. With this cohesive resource, our objective is to empower statisticians and machine learners alike, facilitating a deeper understanding of Bayesian neural networks (BNNs) and promoting their wider practical implementation.

\bibliographystyle{apalike}
\cleardoublepage

\bibliography{references}

\end{document}